\documentclass[10pt,twocolumn,letterpaper]{article}

\usepackage{cvpr}              
\usepackage[accsupp]{axessibility} 
\definecolor{cvprblue}{rgb}{0.21,0.49,0.74}
\usepackage[pagebackref,breaklinks,colorlinks,allcolors=cvprblue]{hyperref}

\usepackage[table]{xcolor}
\usepackage{graphicx}
\usepackage{multirow}
\usepackage{makecell}
\usepackage{pifont}
\usepackage{arydshln}
\newcommand{\cmark}{\ding{51}}%
\newcommand{\xmark}{\ding{55}}%
\usepackage{booktabs}
\usepackage{balance}
\usepackage{subcaption}

\usepackage{times}
\usepackage{amsmath}
\usepackage{amssymb}
\usepackage{enumitem}

\usepackage{array} 
\usepackage{pifont}
\usepackage[normalem]{ulem}

\usepackage{xcolor}
\usepackage{array,tabularx,calc}
\newcolumntype{P}[1]{>{\centering\arraybackslash}p{\dimexpr#1\linewidth\relax}}

\newcommand{\hl}{gray!15}  

\newcommand{\myref}[2]{\hyperref[#1]{#2}}

\def\paperID{} 
\def\confName{CVPR}
\def\confYear{2026}

\title{Learning by Neighbor-Aware Semantics, Deciding by Open-form Flows: Towards Robust Zero-Shot Skeleton Action Recognition}

\author{Yang Chen\textsuperscript{1}, Miaoge Li\textsuperscript{1}, Zhijie Rao\textsuperscript{1}, Deze Zeng\textsuperscript{2}, Song Guo\textsuperscript{3}, Jingcai Guo\textsuperscript{1\thanks{Jingcai Guo is the corresponding author.}}\\
\textsuperscript{1}The Hong Kong Polytechnic University, Hong Kong SAR\\
\textsuperscript{2}China University of Geoscience, China\\
\textsuperscript{3}Hong Kong University of Science and Technology, Hong Kong SAR\\
{\tt jc-jingcai.guo@polyu.edu.hk}
}

\begin{document}
\maketitle
\begin{abstract}
Recognizing unseen skeleton action categories remains highly challenging due to the absence of corresponding skeletal priors. Existing approaches generally follow an ``align-then-classify'' paradigm but face two fundamental issues, \textit{i.e.}, (i) fragile point-to-point alignment arising from imperfect semantics, and (ii) rigid classifiers restricted by static decision boundaries and coarse-grained anchors. To address these issues, we propose a novel method for zero-shot skeleton action recognition, termed \texttt{\textbf{Flora}}, which builds upon \textbf{F}lexib\textbf{L}e neighb\textbf{O}r-aware semantic attunement and open-form dist\textbf{R}ibution-aware flow cl\textbf{A}ssifier. Specifically, we flexibly attune textual semantics by incorporating neighboring inter-class contextual cues to form direction-aware regional semantics, coupled with a cross-modal geometric consistency objective that ensures stable and robust point-to-region alignment. Furthermore, we employ noise-free flow matching to bridge the modality distribution gap between semantic and skeleton latent embeddings, while a condition-free contrastive regularization enhances discriminability, leading to a distribution-aware classifier with fine-grained decision boundaries achieved through token-level velocity predictions. Extensive experiments on three benchmark datasets validate the effectiveness of our method, showing particularly impressive performance even when trained with only 10\% of the seen data. Code is available at~\url{https://github.com/cseeyangchen/Flora}.
\end{abstract}    
\section{Introduction}
\label{sec:introduction}

\begin{figure}[!t]
\centering
\includegraphics[width=0.99\linewidth]{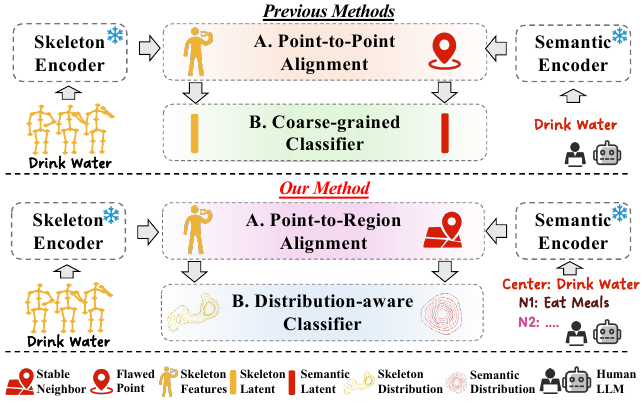}
\caption{Overview of our \texttt{\textbf{Flora}} versus previous methods.}
\vspace{-10pt}
\label{fig:overview}
\end{figure}

Human action recognition has long been a central research topic, driving a wide range of human-centric applications across healthcare~\cite{he2024expert, chen2023stformer}, security~\cite{mishra2024skeletal, sato2023prompt}, and sports~\cite{rao2025towards, hong2021video}. Within this field, zero-shot skeleton action recognition has recently attracted growing attention worldwide. From a modality perspective, the skeleton is inherently data-efficient, privacy-preserving, and illumination-robust compared to other visual inputs. Additionally, the zero-shot setting is highly practical in real-world scenarios, as collecting large-scale behavioral datasets that cover an open-ended range of categories is infeasible, particularly for high-risk and abnormal actions. Consequently, zero-shot skeleton action recognition emerges as an impactful topic for advancing the broader action recognition community.

Basically, zero-shot learning refers to a paradigm in which models are required to recognize categories that are absent during training. In this context, zero-shot skeleton action recognition targets the classification of unseen skeleton actions, in contrast to supervised approaches that are confined to in-domain categories. In practice, two task protocols are commonly considered: (i) the zero-shot learning  (ZSL) setting, where models are evaluated exclusively on unseen actions; (ii) the generalized zero-shot learning (GZSL) setting, where both seen and unseen actions should be recognized. In either case, models are trained using only samples from seen skeleton categories in conjunction with a predefined semantic corpus. This naturally raises the key challenge of how to establish robust relationships between skeletons and semantics. To address this challenge, most prior studies~\cite{jasani2019skeleton,gupta2021syntactically,zhou2023zero,li2023multi, zhu2024part,chen2024fine,li2024sa,xu2025information,kuang2025zero,zhu2025semantic,chen2025neuron,zhu2025prompt,zhou2025zero,wu2025frequency,liu2025beyond,chen2026star++} have widely adopted the so-called ``\textit{align-then-classify}'' paradigm, which first aligns skeletons with semantics and then performs skeleton action classification.

Although a variety of promising techniques have been proposed, two fundamental issues remain insufficiently addressed.
(i) Pursuing desirable semantics for skeletons has emerged as the mainstream paradigm in recent years. However, LLMs-generated descriptive semantics~\cite{zhou2023zero,zhu2024part,chen2024fine,li2023multi,xu2025information,chen2025neuron,wu2025frequency,chen2026star++} inherently deviate due to the lack of explicit skeletal guidance, while parameter-efficient fine-tuned semantics~\cite{zhu2025semantic,zhou2025zero,zhu2025prompt} often overfit to seen skeleton priors. Both approaches result in rigid, flawed, and suboptimal semantic anchors, which in turn cause instability in point-to-point cross-modal alignment. As a result, some action categories may be correctly aligned, while others suffer from collateral misalignment. This process can be likened to navigating with a flawed map that may mislead travelers onto wrong paths. 
(ii) In the classification phase, generative-based methods~\cite{gupta2021syntactically,li2023multi,li2024sa,wu2025frequency} synthesize unseen skeleton features from semantics to train unseen linear classifiers, whereas embedding-based methods~\cite{jasani2019skeleton,zhou2023zero,zhu2024part,chen2024fine,xu2025information,kuang2025zero,zhu2025semantic,chen2025neuron,zhu2025prompt,zhou2025zero,liu2025beyond,chen2026star++} directly rely on cosine similarity for matching. The former inevitably imposes static decision boundaries that cannot adapt to newly emerging categories, while the latter provides better scalability but compresses features into a single vector, leading to information loss and coarse-grained classification. 
\textit{\textbf{Consequently, achieving robust zero-shot skeleton action recognition requires both the calibration of semantics in advance and the development of a flexible and fine-grained classifier.}}

To address the aforementioned issues, we revisit the ``learning'' and ``deciding'' phases of the classical ``\textit{align-then-classify}'' paradigm and introduce \textbf{\texttt{Flora}}, a robust framework for zero-shot skeleton action recognition, as illustrated in Fig.~\ref{fig:overview}. Specifically, in the learning phase, we first locally attune each semantic feature by incorporating adjacent, stable in-context semantics from other categories through similarity-based graph updating. This process produces neighbor-aware contextualized semantics characterized by inherent smoothness and continuity. Building on this foundation, we employ a cross-modal VAE variant with an introduced geometric consistency objective to align the attuned semantics with skeleton features in the latent space, thereby ensuring discriminative point-to-region alignment supported by local neighborhood stability. This shifts the paradigm from blindly groping for an alignment path with unreliable anchors to actively navigating by consulting the overall orientation of nearby stable landmarks, ultimately resulting in more robust alignment. In the deciding phase, inspired by flow matching that transports Gaussian noise into arbitrary distributions, we generalize this concept to model the transformations between learned cross-modal latent distributions. Token-level correspondences between semantic and skeleton embeddings are established via noise-free, condition-free distribution mapping, combined with a contrastive regularization strategy that enables precise velocity-based one-step discrimination. This design allows the learned flow classifier to remain open-form in adapting to new unseen categories and free-form without noise or condition constraints, thereby improving generalization and decision flexibility. Building on the above procedures of learning with neighbor-aware semantics and deciding with noise-free flows, \textbf{\texttt{Flora}} advances the ``\textit{align-then-classify}'' paradigm for zero-shot skeleton action recognition, achieving greater generality, flexibility, and robustness.

The main contributions can be summarized as follows:
\begin{itemize}[leftmargin=2em]
    \item We produce neighbor-aware contextualized semantics and geometric consistency objective, enabling smooth and robust point-to-region cross-modal alignment with directional judgment, effectively alleviating fragile alignment caused by imperfect semantics.
    \item We introduce noise-free and condition-free flows, combined with a contrastive strategy, to realize fine-grained distribution transport across cross-modal latent tokens. This design facilitates a new type of plug-and-play classifier that is flexible, discriminative, and highly generalizable to open-world scenarios.
    \item Extensive experiments show that our method achieves state-of-the-art performance in both ZSL and GZSL settings on NTU-60, NTU-120, and PKU-MMD datasets, with particularly impressive performance under low-shot training conditions in seen categories.
\end{itemize}

\section{Related Work}
\label{sec:relatedwork}

\subsection{Zero-shot Skeleton Action Recognition}
Existing approaches to zero-shot skeleton action recognition can be broadly categorized into two groups: embedding-based methods and generative-based methods.  

\vspace{0.3em}
\noindent\textbf{Embedding-based.} 
These methods project skeleton-semantic pairs into a shared embedding space and perform recognition via cosine similarity.
RelationNet~\cite{jasani2019skeleton} first explored skeleton-semantic relationships by designing deep non-linear metrics. Later, the research focused on enriching semantics via LLMs. SMIE~\cite{zhou2023zero} integrates temporal constraints with global semantics, while PURLS \cite{zhu2024part} and STAR~\cite{chen2024fine} align decomposed skeletons with fine-grained LLMs-generated semantics. Further advances include dual alignment in DVTA~\cite{kuang2025zero} and multi-synonym semantics in InfoCPL~\cite{xu2025information}, extended by Neuron~\cite{chen2025neuron} with multi-turn semantics for step-by-step synergistic alignment. Beyond semantic generation, parameter-efficient fine-tuning (PEFT) has also been explored, such as prompt learning in SCoPLe~\cite{zhu2025semantic}, encoder tuning in PGFA~\cite{zhou2025zero}, and LoRA-based prototype construction in PP-CDL~\cite{zhu2025prompt}. Other efforts include BSZSL~\cite{liu2025beyond}, which introduces RGB cues for auxiliary alignment, and TDSM~\cite{do2024tdsm}, which unifies alignment and classification via text-conditioned denoising diffusion.

\vspace{0.3em}
\noindent\textbf{Generative-based.}
These methods first align skeleton-semantic pairs and then synthesize unseen skeleton features from semantics to train a linear classifier. Most of them~\cite{gupta2021syntactically,li2023multi,li2024sa,wu2025frequency} adopt the cross-modal VAE framework in Sec.~\ref{sec:cross_modal_vae}, \dashuline{where the two modalities are implicitly aligned via a shared Gaussian prior. This implicit alignment, however, neglects the geometric consistency of the cross-modal space, leading to degraded category discriminability.} SynSE~\cite{gupta2021syntactically} introduced it by reconstructing skeletons and semantics bidirectionally. GZSSAR~\cite{li2023multi} extended it with LLM-generated multi-type semantics, and SA-DAVE~\cite{li2024sa} decomposed skeleton features into semantically relevant and irrelevant parts. FS-VAE~\cite{wu2025frequency} incorporates frequency analysis and penalizes mismatched pairs during training.

\vspace{0.3em}
\noindent\textbf{[\textit{Summary}]:}
Unlike prior methods, our method advances this field in two key aspects. First, we incorporate neighboring inter-class contextual semantics in advance and employ the geometric consistency objective to achieve relaxed, stable, robust point-to-region alignment with directional awareness. Second, we introduce a distribution-aware classifier that enables token-level discrimination with improved flexibility, robustness, and generalizability.

\subsection{Cross-Modal Flow Matching}
Flow matching~\cite{lipman2022flow,albergo2022building,liu2022flow,lipman2024flow} has recently emerged as a powerful generative paradigm formulated via ordinary differential equations (ODEs). It enables the conditional synthesis of diverse modalities such as images~\cite{esser2024scaling}, videos~\cite{polyak2024movie}, audio~\cite{vyas2023audiobox}, text~\cite{hu2024flow}, action~\cite{ma2024survey}, and motion~\cite{hu2023motion} from Gaussian noise. Theoretically, the source distribution can be replaced with arbitrary ones, inspiring cross-modal extensions~\cite{liu2025flowing,he2025flowtok,gao2025vita} that eliminate explicit noise injection. CrossFlow~\cite{liu2025flowing} pioneers the direct generation of images from text. FlowTok~\cite{he2025flowtok} generalizes this idea to 1D token representations for efficiency, while VITA~\cite{gao2025vita} extends it to action latents for visuomotor control. However, all these methods still regularize the source distribution toward a Gaussian prior via the KL divergence. More recently, contrastive flow matching~\cite{stoica2025contrastive} improves generation quality by enforcing flow-level separation across various conditions.

\vspace{0.3em}
\noindent\textbf{[\textit{Summary}]:}
To our knowledge, we are the first to extend flow matching to a discriminative setting, specifically tailored for zero-shot recognition. Our framework neither requires approximating the source distribution with a Gaussian prior nor injects any noise during training. Moreover, we advance contrastive flow matching from a noise-driven, conditional formulation to a noise-free and condition-free paradigm, yielding a flexible and discriminative classifier.

\section{Preliminaries}
\label{sec:preliminaries}

\begin{figure*}[t]
\begin{center}
\includegraphics[width=\linewidth]{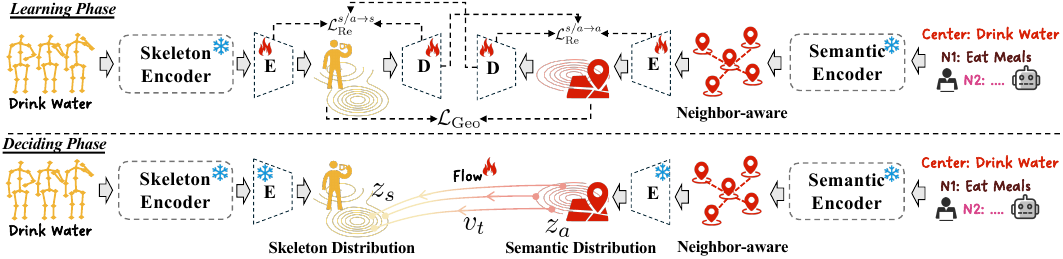} 
\end{center}
\vspace{-15pt}
\caption{The pipeline of our method, including the learning and deciding phases (zoom in for a better view). 
}
\vspace{-10pt}
\label{fig:framework}
\end{figure*}

\noindent\textbf{Cross-modal VAE Alignment.}
\label{sec:cross_modal_vae}
Cross-modal VAE alignment typically builds upon the Variational Autoencoder (VAE), which is optimized as follows:
\begin{equation}
    \mathcal{L}_{\mathrm{VAE}}=\mathbb{E}_{q_{\phi}(z|x)}[\mathrm{log}\,p_{\theta}(x|z)]-\beta D_{KL}(q_{\phi}(z|x)||p_{\theta}(z)),
    \label{eq:vae}
\end{equation}
where the first term denotes the reconstruction error and the second term represents the Kullback-Leibler divergence. Here, $x$ is the input feature, $q_{\phi}(z|x)=\mathcal{N}(\mu, \sigma^2)$ is the encoder that generates the latent representation $z$ via the reparametrization trick~\cite{kingma2013auto}, $p_{\theta}(z)$ is typically a standard Gaussian prior, and $\beta$ balances the KL term~\cite{higgins2017beta}. Previous studies~\cite{gupta2021syntactically,li2023multi,li2024sa,wu2025frequency} adopt a dual-VAE architecture, where two VAEs—one for skeletons and one for semantics—are trained through a cross-reconstruction objective:
\begin{equation}
    \mathcal{L}_{\mathrm{CMR}}
    = \sum_{k\in\{s,a\}} \mathbb{E}_{q_{\phi_k}(z_k|x_k)}[\log p_{\theta_{\bar{k}}}(x_{\bar{k}}|z_k)],
    \label{eq:cmr}
    \vspace{-5pt}
\end{equation}
where $\bar{k}$ denotes the opposite modality of $k$, with $s$ and $a$ representing the skeleton and semantic modalities, respectively. The overall training objective is formulated as:
\begin{equation}
    \mathcal{L}_{\mathrm{CrossVAE}}= \sum_{k\in\{s,a\}}\mathcal{L}_{\mathrm{VAE}}^{k} + \mathcal{L}_{\mathrm{CMR}}.
    \vspace{-5pt}
\end{equation}
Obviously, it contains three parts: (i) intra-reconstruction (the first term in Eq.~\ref{eq:vae}); (ii) cross-reconstruction (Eq.~\ref{eq:cmr}); and (iii) latent regularization (the second term in Eq.~\ref{eq:vae}). 

\vspace{0.5em}
\noindent\textbf{Flow Matching.}
\label{sec:flow-matching}
Flow matching aims to learn a velocity field $v_{\theta}$, parameterized by a neural network $\theta$, whose flow $\phi_t$ defines a probability path $p_t$ that transforms samples $z_0\sim p_0$ into corresponding samples $z_1\sim p_1$. The time-dependent flow $\phi_t$ is governed by the following ordinary differential equation (ODE):
\begin{equation}
    \frac{\mathrm{d} }{\mathrm{d} t} \phi_{t}(z) =v_{\theta}(z_t, t),\;\;\;\phi_{0}(z)=z_0,
\end{equation}
where $z_t=\phi_{t}(z)\sim p_t$ represents intermediate samples at continuous time $t\in[0,1]$. For computational efficiency, the probability path between the source and the target distributions is typically linearly interpolated at time $t$~\cite{lipman2022flow,albergo2022building,liu2022flow}:
\begin{equation}
    z_t=[1-(1-\sigma_{\mathrm{min}})t]z_0+tz_1,\;\;\;\sigma_{\mathrm{min}}=10^{-5}.
    \label{eq:interpolated}
\end{equation}
Accordingly, the ground-truth velocity field is given by:
\begin{equation}
    v^* = \frac{\mathrm{d} z_t}{\mathrm{d} t}=z_1-(1-\sigma_{\mathrm{min}})z_0.
    \label{eq:ground-truth}
\end{equation}
The neural velocity field $v_{\theta}$ is then optimized by minimizing MSE between the predicted and ground-truth velocities. Later, in generative tasks such as image synthesis, the target image $\hat{z}_1$ can be generated by integrating during inference.
Typically, these generative methods require that $z_0$ follows a standard Gaussian~\cite{esser2024scaling,polyak2024movie,vyas2023audiobox,hu2024flow,ma2024survey,hu2023motion} or a pseudo-Gaussian approximation~\cite{liu2025flowing,he2025flowtok,gao2025vita}. 

\section{Method}
\label{sec:method}

\subsection{Problem Formulation}
\label{sec:formulation}
Let $\mathcal{D}=\{(\mathbf{X}_{i}, y_{i})\}_{i=1}^{N}$ be the skeleton dataset, where $N$ is the number of samples and $\mathcal{Y}$ denotes the set of action categories with $|\mathcal{Y}|$ classes.  
Each category $y \in \mathcal{Y}$ is associated with a semantic description $a_y$, forming the semantic set $\mathcal{A}=\{a_y\}_{y\in\mathcal{Y}}$.  
Each skeleton sequence $\mathbf{X}_{i}\in \mathbb{R}^{3\times T\times V\times M }$ corresponds to an action label $y_{i}\in\mathcal{Y}$ and its semantic $a_{y_i}$, where 3 represents 3D joint coordinates, $T$ is the number of frames, $V$ is the number of joints, and $M$ indicates the number of human subjects. The dataset is organized into three subsets: a training set $\mathcal{D}_{\text{tr}}^{s}$ containing $\left | \mathcal{Y}^{s}  \right |$ seen categories, a test set $\mathcal{D}_{\text{te}}^{s}$ comprising the same seen categories, and another test set $\mathcal{D}_{\text{te}}^{u}$ consisting of $\left | \mathcal{Y}^{u}  \right |$ unseen categories.  
By definition, $\mathcal{Y} = \mathcal{Y}^{s} \cup \mathcal{Y}^{u}$ and $\mathcal{Y}^{s} \cap \mathcal{Y}^{u}=\varnothing$.  
During training, only $\mathcal{D}_{\text{tr}}^{s}$ is utilized.  
At inference time, the ZSL setting evaluates the model on $\mathcal{D}_{\text{te}}^{u}$, while the GZSL setting evaluates it on both seen and unseen categories using $\mathcal{D}_{\text{te}}^{s} \cup \mathcal{D}_{\text{te}}^{u}$.  
For conciseness, we omit the superscripts denoting seen ($s$) and unseen ($u$) categories in the subsequent sections.

\subsection{Neighbor-aware Semantic Learning}
\label{sec:learning}
\noindent\textbf{Neighbor Semantic Attunement.}
Instead of passively correcting flawed semantics through complex model architectures during the alignment phase, it is more effective to proactively renew them in advance. 
Specifically, for each text semantic $a_y \in \mathcal{A}$, we first extract its feature $\mathbf{F}_{a_y} \in \mathbb{R}^{M_a \times d_a}$ using a pre-trained text encoder $\Psi_\mathbf{text}(\cdot)$, where $M_a$ denotes the number of tokens and $d_a$ is the feature dimension.
Although the LLM-generated semantic $a_y$ may deviate slightly, much like a landmark whose marked position on a map is slightly off, its relative location with respect to surrounding landmarks remains consistent. Hence, we empirically assume that neighborhood relations among semantics are reliable and incorporate these relationships to refine each semantic representation. 
Formally, for a given semantic $a_y$, we define $\mathcal{\hat{A}}_y=\{a_{y'}|y'\in \mathcal{Y}, y' \ne y\}$ as the set of semantics from the remaining categories. Then, we compute the pairwise cosine similarity scores between the semantic feature $F_{a_y}$ and all others $\{\mathbf{F}_{a_{y'}} \mid a_{y'} \in \mathcal{\hat{A}}_y\}$, and select the top-$k$ neighbors via a top-$k$ argmax:
\begin{equation}
    \mathcal{T}_{k}(\mathbf{F}_{a_y})
    = \arg\max_{a_{y'} \in \hat{\mathcal{A}}_{y}}^{(k)}
      \frac{\mathbf{f}_{a_y} \cdot \mathbf{f}_{a_{y'}}}
           {\|\mathbf{f}_{a_y}\| \, \|\mathbf{f}_{a_{y'}}\|},
    \vspace{-5pt}
\end{equation}
where $\mathbf{f}_{a_y} = \rho(\mathbf{F}_{a_y})$ and $\mathbf{f}_{a_{y'}} = \rho(\mathbf{F}_{a_{y'}})$ are the pooled features obtained by applying pooling operation $\rho(\cdot)$ across the token dimension. Subsequently, the selected neighbors $\mathcal{T}_k(\mathbf{F}_{a_y})$ are treated as graph nodes and aggregated them to transform the potentially biased semantic anchor $\mathbf{F}_{a_y}$ into the stable, contextualized, and neighbor-aware semantic representation $\mathbf{O}_{a_y}$ as follows:
\begin{equation}
    \mathbf{O}_{a_y}=\mathbf{F}_{a_y}+ \frac{\tau}{k} \cdot \sum_{\mathbf{F}_{i} \in \mathcal{T}_k(\mathbf{F}_{a_y})}w_i \cdot \mathbf{F}_{i},
    \vspace{-5pt}
\end{equation}
where the coefficient $\tau$ prevents over-smoothing and preserves category-level discriminability. $w_i$ is the corresponding similarity score to control the contribution of neighbors. 

\vspace{0.3em}
\noindent\textbf{Geometric Consistency Alignment.}
\label{sec:distrubition_alignment}
Building upon the neighbor-aware semantics $\mathbf{O}_{a_y}$, we also extract skeleton feature $\mathbf{F}_s\in \mathbb{R}^{1\times d_s}$ using a pre-trained skeleton encoder $\Phi_{\mathbf{skeleton}}(\cdot)$, where $d_s$ is the skeleton feature dimension. To maintain token-level consistency during alignment and preserve fine-grained semantic details, the skeleton feature $\mathbf{F}_s$ is expanded along the token dimension, yielding $\mathbf{\hat{F}}_s \in \mathbb{R}^{M_a \times d_s}$. For each paired skeleton feature $\mathbf{\hat{F}}_s$ and semantic feature $\mathbf{O}_{a_y}$ (denoted as $x_s$ and $x_a$ for brevity), we produce their distributions $z_s \sim \mathcal{N}_s(\mu_s, \sigma_s)$ and $z_a \sim \mathcal{N}_a(\mu_a, \sigma_a)$ via the encoder of respective VAE with the reparameterization trick. Following prior works~\cite{gupta2021syntactically,li2023multi,li2024sa,wu2025frequency}, we retain both intra- and cross-reconstruction objectives, unified as:
\begin{equation}
\mathcal{L}_{\mathrm{Re}}
= \!\!\sum_{k\in\{s,a\}}\!\!
\mathbb{E}_{q_{\phi_k}}[\log p_{\theta_k}(x_k|z_k)+\log p_{\theta_{\bar{k}}}(x_{\bar{k}}|z_k)],
\vspace{-5pt}
\end{equation}
where $\bar{k}$ denotes the opposite modality of $k$. 
Later, we directly align the geometric structure of two modalities in the latent space rather than regularize them toward a standard Gaussian prior (the second term in Eq.~\ref{eq:vae}) as follows:
\begin{equation}
\mathcal{L}_{\mathrm{Geo}}
=   \| \mu_s-\mu_a \|_2^2 + \|\sigma_s^{2}-\sigma_a^{2}\|_2^2.
\vspace{-5pt}
\end{equation}
It encourages the skeleton and semantic distributions to align closely, narrowing the modality gap while preserving inter-class separability and ensuring coherent cross-modal correspondence. Furthermore, the regional semantics provide stable point-to-region support, effectively mitigating alignment instability and enhancing the generalization of skeleton representations to unseen actions.
The overall alignment objective is then summarized as:
\begin{equation}
\mathcal{L}_{\mathrm{Align}}=\mathcal{L}_{\mathrm{Re}}+ \lambda_{\mathrm{Align}} \cdot \mathcal{L}_{\mathrm{Geo}}
\label{eq:phase1}
\vspace{-5pt}
\end{equation}
where $\lambda_{\mathrm{Align}}$ controls the trade-off between reconstruction fidelity and distribution consistency. 

\subsection{Open-form Flow Deciding}
\label{sec:deciding}
\noindent\textbf{Noise-free Flow Mapping.}
Although the cross-modal alignment has been established, an inherent modality gap still remains between $z_s$ and $z_a$~\cite{zhang2024connect}. This theoretical evidence provides solid motivation for introducing a potential distribution transport bridge between the source $\mathcal{N}_a$ and target $\mathcal{N}_s$ via vanilla flow matching. Importantly, $\mathcal{N}_a$ is not Gaussian-approximated nor noise-injected, resulting in a purely noise-free formulation. Then, we interpolate intermediate samples $z_t \in \mathbb{R}^{M_a \times d}$ via Eq.~\ref{eq:interpolated} and obtain the ground-truth velocity $v^{*}$ via Eq.~\ref{eq:ground-truth} to optimize velocity field:
\begin{equation}
    \mathcal{L}_{\mathrm{Flow}}=\mathbb{E}_{t,z_s,z_a} \|v_{\theta}(z_t, t)-v^* \|^2_2,
    \label{eq:fm}
    \vspace{-5pt}
\end{equation}
where $v_{\theta}(z_t, t)$ denotes the predicted velocity. 

\vspace{0.3em}
\noindent\textbf{Condition-free Contrastive Deciding.}
Since the latent distributions of both the source and target vary across categories, \textit{i.e.}, showing inter-class embedding separability, the corresponding transport paths between $\mathcal{N}_a$ and $\mathcal{N}_s$ are also discriminative (in Fig.~\ref{fig:flow_distribution_with_velocity}). This motivates us to directly compare the token-level velocity fields predicted for different semantics to perform skeleton recognition, which is inherently fine-grained, information-preserving, and distribution-aware. Moreover, the motion and semantically enriched distributions enable condition-free transport, contrasting with previous noise-driven conditional flow models. To further enhance the discriminative capability of the learned velocity field for classification, we employ the contrastive regularization term~\cite{stoica2025contrastive} into Eq.~\ref{eq:fm} to shape an overall deciding objective as follows: 
\begin{equation}
    \mathcal{L}_{\mathrm{ConFlow}}=\mathbb{E}_{t,z_s,z_a}\begin{bmatrix}
 \|v_{\theta}(z_t, t)-v^* \|^2_2 \\
- \lambda_{\mathrm{Flow}} \|v_{\theta}(z_t, t)-\hat{v}^* \|^2_2
\end{bmatrix},
\label{eq:phase2}
\vspace{-5pt}
\end{equation}
where $\hat{v}^*$ is the ground-truth velocity computed using skeleton-semantic pairs from other categories, and $\lambda_{\mathrm{Flow}}$ controls the strength of contrastive regularization. This design allows the flow-based classifier to function in an open-form manner, achieving noise-free, condition-free, and boundary-free decision-making with plug-and-play efficiency and fine-grained discriminability.

\subsection{Training \& Prediction}
\label{sec:prediction}
\noindent\textbf{Training Pipeline.}
We first optimize Eq.\ref{eq:phase1} independently, and then freeze its parameters to train Eq.\ref{eq:phase2} separately.

\vspace{0.3em}
\noindent\textbf{ZSL Prediction.}
For each unseen skeleton latent embedding $z_s$ and its candidate semantic set $\{z_{a_y}|, y \in \mathcal{Y}^u\}$, we compute the ground-truth velocity $v_y^*$  for each to-be-matched pair using Eq.~\ref{eq:ground-truth}. The interpolated latent embedding $z_t^y$ at time $t$ is then fed into the flow classifier to produce the one-step predicted velocity $v_{\theta}(z_t^y, t)$. Then, we define the velocity error as $\varepsilon_y=\| v_{\theta}(z_t^y, t)- v_{y}^*  \|_2$ and select the minimal as the classification result:
\begin{equation}
    \hat{y}=\arg \min_{y \in \mathcal{Y}^{u}} \varepsilon_y.
    \vspace{-5pt}
\end{equation}
Compared with the static classifiers in~\cite{gupta2021syntactically,li2023multi,li2024sa,wu2025frequency}, this prediction pipeline is easily extendable to new categories without retraining. Meanwhile, it retains token-level fine-grained information for recognition, offering higher representational fidelity than the vector-compressed cosine similarity classifiers in~\cite{jasani2019skeleton,zhou2023zero,zhu2024part,chen2024fine,zhu2025semantic,chen2025neuron}.

\vspace{0.3em}
\noindent\textbf{GZSL Prediction.}
For each skeleton latent embedding $z_s$, we first compute the minimal velocity error $\delta_{\mathcal{Y}^s}$ and $\delta_{\mathcal{Y}^u}$ over seen and unseen categories, respectively, where $\delta_{\mathcal{Y}^s/\mathcal{Y}^u} = \min_{y \in \mathcal{Y}^s/\mathcal{Y}^u} \varepsilon_y$. Their ratio indicates whether the input is more likely to belong to the seen or unseen domain: a lower ratio suggests a higher likelihood for the seen domain due to training on seen categories, and vice versa. We then set a threshold $\gamma$ to determine the category domain before recognition. Once finished, we only recognize the skeleton in the respective domains. The unified prediction formulation is expressed as:
\begin{equation}
    \hat{y}=\arg\min_{y\in\mathcal{Y}}\big[\varepsilon_y+\alpha\cdot\mathbb{I}[(y\in\mathcal{Y}^s)\oplus(\frac{\delta_{\mathcal{Y}^s}}{\delta_{\mathcal{Y}^u}}  \le\gamma)]\big],
    \vspace{-5pt}
\end{equation}
where  $\alpha \gg 1$ is a large penalty coefficient and $\oplus$ is the Exclusive OR (XOR) operator. 

\section{Experiments}
\label{sec:experiment}


\begin{table*}[!h]
\caption{Performance comparison on NTU-60 (Xsub) and NTU-120 (Xsub). The best and the second-best results are marked in \textcolor{red}{\textbf{Red}} and \textcolor{blue}{Blue}, respectively. $^{\color{Purple}\boldsymbol{\dagger}}$ denotes methods using SynSE-based~\cite{gupta2021syntactically} Shift-GCN features, while others use STAR-based~\cite{chen2024fine} ones. Both are trained in the same manner~\cite{cheng2020skeleton}, but differ slightly and were inconsistently used in prior works, so we report both for completeness and fairness. $^{\color{Orange}\boldsymbol{\ddagger}}$ indicates two-stream fusion; others are single-stream. Results on Xview and Xset are reported in the \textcolor{magenta}{Appendix}. }
\vspace{-5pt}
\label{table:ntu60_ntu120_xsub}
\centering
\scalebox{0.75}{
\renewcommand{\arraystretch}{1.0}
\begin{tabular}{lccccccccccccccccc}
\toprule
\multirow{5}{*}{Method} & \multirow{5}{*}{Venue} & \multicolumn{8}{c}{NTU RGB+D 60 (Xsub)} & \multicolumn{8}{c}{NTU RGB+D 120 (Xsub)}\\
\cmidrule(r){3-10} \cmidrule(l){11-18}
& & \multicolumn{4}{c}{55/5 Split} & \multicolumn{4}{c}{48/12 Split} & \multicolumn{4}{c}{110/10 Split} & \multicolumn{4}{c}{96/24 Split}\\
\cmidrule(r){3-6} \cmidrule(lr){7-10} \cmidrule(lr){11-14} \cmidrule(l){15-18}
& & ZSL & \multicolumn{3}{c}{GZSL} &  ZSL & \multicolumn{3}{c}{GZSL} &  ZSL & \multicolumn{3}{c}{GZSL} &  ZSL & \multicolumn{3}{c}{GZSL} \\
\cmidrule(r){3-3} \cmidrule(lr){4-6} \cmidrule(lr){7-7} \cmidrule(lr){8-10} \cmidrule(lr){11-11} \cmidrule(lr){12-14} \cmidrule(lr){15-15} \cmidrule(l){16-18}
& & $\mathcal{A}cc$ & $\mathcal{S}$ & $\mathcal{U}$ & $\mathcal{H}$ & $\mathcal{A}cc$ & $\mathcal{S}$ & $\mathcal{U}$ & $\mathcal{H}$ & $\mathcal{A}cc$ & $\mathcal{S}$ & $\mathcal{U}$ & $\mathcal{H}$ & $\mathcal{A}cc$ & $\mathcal{S}$ & $\mathcal{U}$ & $\mathcal{H}$ \\
\midrule
ReViSE~\cite{hubert2017learning} &  ICCV 2017 & 69.5 & 40.8 & 50.2 & 45.0 & 24.0 & 21.8 & 14.8 & 17.6 & 19.8 & 0.6 & 14.5 & 1.1 & 8.5 & 3.4 & 1.5 & 2.1  \\
JPoSE~\cite{wray2019fine} &  ICCV 2019 & 73.7 & 66.5 & 53.5 & 59.3 & 27.5 & 28.6 & 18.7 & 22.6 & 57.3 & 53.6 & 11.6 & 19.1 & 38.1 & 41.0 & 3.8 & 6.9  \\
CADA-VAE~\cite{schonfeld2019generalized} & CVPR 2019 & 76.9 & 56.1 & 56.0 & 56.0 & 32.1 & 50.4 & 25.0 & 33.4 & 52.5 & 50.2 & 43.9 & 46.8 & 38.7 & 48.3 & 27.5 & 35.1 \\
SynSE~\cite{gupta2021syntactically} & ICIP 2021 & 71.9 & 51.3 & 47.4 & 49.2 & 31.3 & 44.1 & 22.9 & 30.1 & 52.4 & 57.3 & 43.2 & 49.5 & 41.9 & 48.1 & 32.9 & 39.1 \\
GZSSAR$^{\color{Purple}\boldsymbol{\dagger}}$~\cite{li2023multi} & ICIG 2023 & 83.6 & 71.7 & 66.2 & 68.8 & 49.2 & 58.8 & 40.0 & 47.6 & 71.2 & 46.8 & 68.3 & 55.6 & 59.7 & 56.8 & 48.6 & 52.4 \\
SMIE~\cite{zhou2023zero} & ACMMM 2023 &77.9 & - & - & - & 41.5 & - & - & - & 61.3 & - & - & - & 42.3 & - & - & -\\
PURLS~\cite{zhu2024part} & CVPR 2024 &79.2 & - & - & - & 41.0 & - & - & - & 72.0 & - & - & - & 52.0 & - & - & -\\
SA-DAVE~\cite{li2024sa} & ECCV 2024 & 82.4 & 62.8 & 70.8 & 66.3 & 41.4 & 50.2 & 36.9 & 42.6 & 68.8 & 61.1 & 59.8 & 60.4 & 46.1 & 58.8 & 35.8 & 44.5 \\
STAR~\cite{chen2024fine} & ACMMM 2024 & 81.4 & 69.0 & 69.9 & 69.4 & 45.1 & 62.7 & 37.0 & 46.6 & 63.3 & 59.9 & 52.7 & 56.1 & 44.3 & 51.2 & 36.9 & 42.9  \\
STAR++~\cite{chen2026star++} & TCSVT 2026 & 84.4 & 61.1 & 73.6 & 66.8 & 49.5 & 58.2 & 40.4 & 47.7 & 72.0 & 59.0 & 55.4 & 57.2 & 53.5 & 52.8 & 45.2 & 48.7 \\
DVTA$^{\color{Purple}\boldsymbol{\dagger}}$~\cite{kuang2025zero} & PR 2025 & 79.3 & - & - & - & 44.1 & - & - & - & \textcolor{blue}{74.9} & - & - & - & 51.8 & - & - & - \\
InfoCPL$^{\color{Purple}\boldsymbol{\dagger}}$~\cite{xu2025information} & TMM 2025 & 85.9 & - & - & - & 53.3 & - & - & - & 74.8 & - & - & - & 60.1 & - & - & - \\
ScoPLe$^{\color{Purple}\boldsymbol{\dagger}}$~\cite{zhu2025semantic} & CVPR 2025 & 84.1 & 69.6 & 71.9 & 70.8 & 53.0 & 54.5 & \textcolor{red}{\textbf{61.8}} & 57.9 &  74.5 & 63.5 & 61.1 & 62.3 & 52.2 & 53.3 & \textcolor{blue}{51.2} & 52.2 \\
Neuron$^{\color{Orange}\boldsymbol{\ddagger}}$~\cite{chen2025neuron} & CVPR 2025 & \textcolor{blue}{86.9} & 69.1 & 73.8 & 71.4 & \textcolor{blue}{62.7} & \textcolor{blue}{61.6} & 56.8 & \textcolor{blue}{59.1} & 71.5 & \textcolor{red}{\textbf{67.6}} & 59.5 & 63.3 & 57.1 & \textcolor{red}{\textbf{67.5}} & 44.4 & \textcolor{blue}{53.6} \\
FS-VAE$^{\color{Purple}\boldsymbol{\dagger}}$~\cite{wu2025frequency} & ICCV 2025 & \textcolor{red}{\textbf{86.9}} & \textcolor{red}{\textbf{77.0}} & \textcolor{blue}{74.5} & \textcolor{blue}{75.7} & 57.2 & 56.2 & 48.6 & 52.1 & 74.4 & 59.2 & \textcolor{red}{\textbf{67.9}} & \textcolor{blue}{63.3} & 62.5 & \textcolor{blue}{57.8} & \textcolor{red}{\textbf{51.9}} & \textcolor{red}{\textbf{54.7}} \\
TDSM$^{\color{Purple}\boldsymbol{\dagger}}$~\cite{do2024tdsm} &  ICCV 2025 & 86.5 & - & - & - & 56.0 & - & - & - & 74.2 & - & - & - & \textcolor{blue}{65.1} & - & - & -  \\
\hdashline
\rowcolor{yellow!10} \textbf{\texttt{Flora} (Ours)} & This work & 85.8 & 77.7 & 75.6 & 76.6 & 61.5 & 66.9 & 49.0 & 56.6 & 80.7 & 59.8 & 70.5 & 64.7 & 64.1 & 53.7 & 52.2 & 52.9 \\
\rowcolor{yellow!10} \textbf{\texttt{Flora} (Ours)}$^{\color{Purple}\boldsymbol{\dagger}}$ & This work & 86.3 & \textcolor{blue}{75.9} & \textcolor{red}{\textbf{78.8}} & \textcolor{red}{\textbf{77.4}} & \textcolor{red}{\textbf{65.3}} & \textcolor{red}{\textbf{63.7}} & \textcolor{blue}{57.5} & \textcolor{red}{\textbf{60.5}} & \textcolor{red}{\textbf{79.6}} & \textcolor{blue}{66.2} & \textcolor{blue}{66.0} & \textcolor{red}{\textbf{66.1}} & \textcolor{red}{\textbf{66.4}} & 55.9 & 50.7 & 53.2 \\
\bottomrule
\end{tabular}}
\vspace{-5pt}
\end{table*}


To evaluate the effectiveness of \texttt{\textbf{Flora}}, we conduct comprehensive experiments across three mainstream datasets: NTU-60 ~\cite{shahroudy2016ntu}, NTU-120~\cite{liu2019ntu}, and PKU-MMD~\cite{liu2017pku}. The dataset introduction is illustrated in \textcolor{magenta}{Appendix~\ref{app:datasets}}. For more experimental details, results, and analyses beyond the main body of the paper, we encourage readers to the \textcolor{magenta}{Appendix}.


\subsection{Experiment Settings}
\label{sec:experiment_settings}
Our work follows the seen/unseen category split protocols established in prior studies~\cite{gupta2021syntactically, li2024sa}, including basic split protocols~\cite{gupta2021syntactically,chen2024fine}, random split protocols~\cite{li2024sa, chen2024fine, zhou2023zero}, and challenging split protocols~\cite{zhu2024part}. 
The split details are provided in \textcolor{magenta}{Appendix~\ref{app:dataset_splits}}. For evaluation, we report the Top-1 accuracy $\mathcal{A}cc=\frac{1}{N}\sum_{i=1}^{N} \mathbb{I}[y_i=\hat{y}]$ on the $\mathcal{D}_{\text{te}}^{u}$ in the ZSL setting. In the GZSL setting, we report the accuracy on seen classes ($\mathcal{S}$) using $\mathcal{D}_{\text{te}}^{s}$, the accuracy on unseen classes ($\mathcal{U}$) using $\mathcal{D}_{\text{te}}^{u}$, and their harmonic mean accuracy ($\mathcal{H}=(2\times\mathcal{S}\times\mathcal{U})/(\mathcal{S}+\mathcal{U})$). Additional implementation details are provided in \textcolor{magenta}{Appendix~\ref{app:implementation_details}}.


\subsection{Performance Comparison}


\noindent\textbf{Basic Split Benchmark Evaluation I}.
Table~\ref{table:ntu60_ntu120_xsub} presents a comparison between our method and other approaches on the Xsub benchmarks of NTU-60 and NTU-120, using both SynSE-extracted 4s-Shift-GCN and STAR-extracted 1s-Shift-GCN skeleton features. Across both settings, our method consistently achieves competitive performance in both ZSL and GZSL scenarios, with particularly strong results on NTU-60 (48/12 split) and NTU-120 (110/10 split). 
Additional evaluations on the Xview and Xset benchmarks are provided in the \textcolor{magenta}{Appendix~\ref{app:addtional_sota}}.


\vspace{0.3em}
\noindent\textbf{Low-shot Training Sample Evaluation}.
We further evaluate our method under the low-shot setting with SynSE-based~\cite{gupta2021syntactically} Shift-GCN features, where only a small fraction of training samples is available for each seen category. As shown in Table~\ref{table:low_shot_training_part}, our approach achieves competitive performance even with only 1\% of the training data, surpassing all prior methods a lot and showing strong generalization with extremely limited seen skeleton priors. The complete results are provided in \textcolor{magenta}{Appendix~\ref{app:addtional_sota}}.

\begin{table}[!h]
\vspace{-10pt}
\caption{ZSL Comparison under low-shot training.}
\label{table:low_shot_training_part}
\vspace{-7pt}
\centering
\scalebox{0.66}{
\renewcommand{\arraystretch}{1.0}
\begin{tabular}{lcccccccc}
\toprule
\multirow{3}{*}{Method} & \multicolumn{4}{c}{NTU-60} & \multicolumn{4}{c}{NTU-120} \\
\cmidrule(r){2-5} \cmidrule(lr){6-9}
 & \multicolumn{2}{c}{55/5 (Xsub)} & \multicolumn{2}{c}{48/12 (Xsub)} & \multicolumn{2}{c}{110/10 (Xsub)} & \multicolumn{2}{c}{96/24 (Xsub)}\\
\cmidrule(r){2-3} \cmidrule(lr){4-5} \cmidrule(r){6-7} \cmidrule(lr){8-9}
& 1\% & 10\% & 1\% & 10\% & 1\% & 10\% & 1\% & 10\%  \\
\midrule
ReViSE~\cite{hubert2017learning} & 51.0 & 58.0 & 9.8 & 15.6 & 14.8 & 23.6 & 5.2 & 7.8 \\
JPoSE~\cite{wray2019fine} & 33.0 & 62.1 & 23.8 & 28.3 & 15.6 & 49.9 & 8.6 & 33.9 \\
CADA-VAE~\cite{schonfeld2019generalized} & 76.6 & 76.9 & 24.3 & 27.6 & 29.9 & 39.1 & 25.4 & 25.0 \\
SynSE~\cite{gupta2021syntactically} & 44.3 & 42.8 & 18.6 & 17.3 & 56.0 & 56.0 & 24.1 & 26.1 \\
SMIE~\cite{zhou2023zero} & 43.8 & 76.9 & 29.3 & 38.1 & 36.0 & 58.1 & 13.9 & 34.4  \\
SA-DAVE~\cite{li2024sa} & 21.1 & 60.4 & 18.7 & 20.0 & 14.4 & 40.9 & 9.5 & 21.9 \\
STAR~\cite{chen2024fine} & 40.6 & 77.0 & 11.6 & 35.3 & 18.9 & 46.8 & 8.6 & 33.0  \\
Neuron~\cite{chen2025neuron} & 47.7 & 79.4 & 20.7 & 45.3 & 28.8 & 62.8 & 10.2 & 33.5 \\
FS-VAE~\cite{wu2025frequency} & \textcolor{blue}{79.3} & 79.4 & \textcolor{blue}{38.0} & 38.7 & \textcolor{blue}{72.7} & \textcolor{blue}{69.6} & \textcolor{blue}{50.2} & 47.9 \\
TDSM~\cite{do2024tdsm} & 78.5 & \textcolor{blue}{82.3} & 32.1 & \textcolor{blue}{52.4} & 63.3 & 66.3 & 43.9 & \textcolor{blue}{55.1} \\
\hdashline
\rowcolor{yellow!10} \textbf{\texttt{Flora} (Ours)} & \textcolor{red}{\textbf{82.8}} & \textcolor{red}{\textbf{85.6}} & \textcolor{red}{\textbf{46.5}} & \textcolor{red}{\textbf{56.1}} & \textcolor{red}{\textbf{77.4}} & \textcolor{red}{\textbf{78.1}} & \textcolor{red}{\textbf{58.0}} & \textcolor{red}{\textbf{65.9}} \\
\bottomrule
\end{tabular}}
\vspace{-10pt}
\end{table}


\vspace{0.3em}
\noindent\textbf{Random Split Benchmark Evaluation I}.
We also evaluate our method on the random split benchmark following the protocol in~\cite{li2024sa}. Each dataset has three randomly selected seen-unseen splits, and the skeleton features are extracted using ST-GCN. The average results across the splits are reported in Table~\ref{table:random_split_sadave_based}. As observed, our method consistently outperforms all competitors across the three datasets, particularly achieving significant gains under the GZSL metric.

\begin{table}[h!]
\caption{Average performance on three random seen–unseen splits (SA-DAVE~\cite{li2024sa}, ST-GCN features). STAR-based~\cite{chen2024fine} results with Shift-GCN features are in the \textcolor{magenta}{Appendix}.}
\label{table:random_split_sadave_based}
\vspace{-5pt}
\centering
\scalebox{0.75}{
\renewcommand{\arraystretch}{1.0}
\begin{tabular}{lcccccc}
\toprule
\multirow{3}{*}{Method} & \multicolumn{2}{c}{NTU-60} & \multicolumn{2}{c}{NTU-120} & \multicolumn{2}{c}{PKU-MMD I}\\
 & \multicolumn{2}{c}{55/5 (Xsub)} & \multicolumn{2}{c}{110/10 (Xsub)} & \multicolumn{2}{c}{46/5 (Xsub)}\\
\cmidrule(r){2-3} \cmidrule(lr){4-5} \cmidrule(l){6-7}
& ZSL & GZSL & ZSL & GZSL & ZSL & GZSL \\
\midrule
ReViSE~\cite{hubert2017learning} & 60.9 & 60.3 & 44.9 & 40.3 & 59.3 & 49.8 \\
JPoSE~\cite{wray2019fine} & 59.4 & 60.1 & 46.7 & 43.7 & 57.2 & 51.6 \\
CADA-VAE~\cite{schonfeld2019generalized} & 61.8 & 66.4 & 45.2 & 45.6 & 60.7 & 45.8 \\
SynSE~\cite{gupta2021syntactically} & 64.2 & 67.5 & 47.3 & 43.5 & 60.8 & 49.5 \\
SMIE~\cite{zhou2023zero} & 65.1 & - & 46.4 & - & 60.8 & - \\
SA-DAVE~\cite{li2024sa} & 84.2 & 75.3 & 50.7 & 47.5 & 66.5 & 54.7 \\
SCoPLe~\cite{zhu2025semantic} & 83.7 & \textcolor{blue}{77.7} & 53.3 & \textcolor{blue}{54.1} & \textcolor{blue}{71.4} & 54.9 \\
TDSM~\cite{do2024tdsm} & \textcolor{red}{\textbf{88.9}} & - & \textcolor{blue}{69.5} & - & 70.8 & - \\
FS-VAE~\cite{wu2025frequency} & - & - & - & - & 71.2 & \textcolor{blue}{59.0} \\
\hdashline
\rowcolor{yellow!10} \textbf{\texttt{Flora} (Ours)} & \textcolor{blue}{88.6} & \textcolor{red}{\textbf{80.2}} & \textcolor{red}{\textbf{71.2}} & \textcolor{red}{\textbf{63.0}} & \textcolor{red}{\textbf{71.6}} & \textcolor{red}{\textbf{59.5}} \\
\bottomrule
\end{tabular}}
\vspace{-10pt}
\end{table}


\begin{figure*}[!h]
\centering
\includegraphics[width=0.99\linewidth]{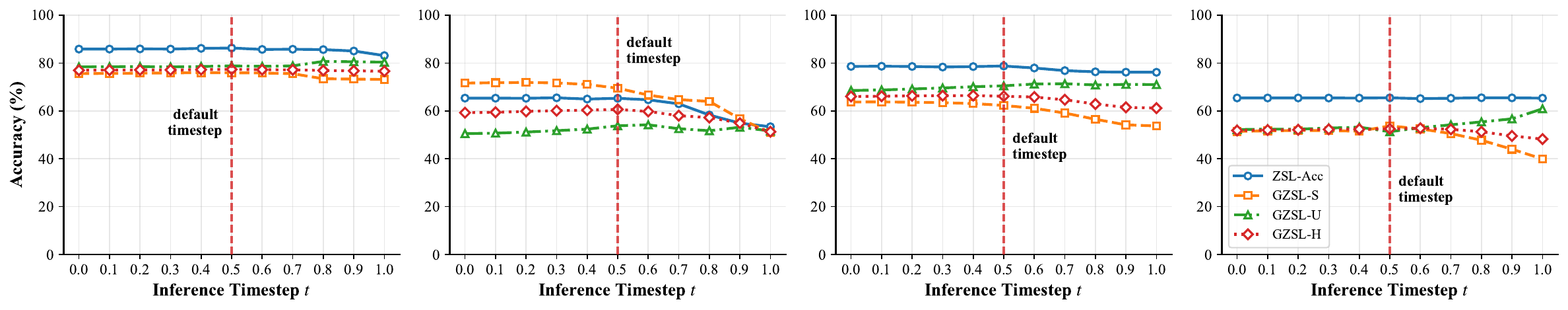} \\
\vspace{-5pt}
\begin{minipage}{0.99\linewidth}
\centering
\begin{tabular}{P{0.26}P{0.21}P{0.22}P{0.21}}
\footnotesize{(a) NTU-60 (55/5)} & 
\footnotesize{(b) NTU-60 (48/12)}  & 
\footnotesize{(c) NTU-120 (110/10)} & 
\footnotesize{(d) NTU-120 (96/24)} \\
\end{tabular}
\end{minipage}
\vspace{-7pt}
\caption{Performance comparison on NTU-60 and NTU-120 with different timestep selection $t$ in the inference phase.}
\label{fig:timestep}
\vspace{-15pt}
\end{figure*}


\subsection{Ablation Studies}


\noindent \textbf{Influence of Components in the Learning Phase}.
To assess component contributions, we remove modules from the learning phase while keeping the deciding phase fixed. As shown in Table~\ref{table:learning_component_analysis}, the geometric consistency objective plays a key role, significantly improving ZSL and GZSL performance, especially under the NTU-60 (48/12) split. Semantic attunement further enhances results, and their combination yields stable cross-modal point-to-region alignment.

\begin{table}[!h]
\caption{Analysis of different components in the learning phase.}
\label{table:learning_component_analysis}
\vspace{-5pt}
\centering
\scalebox{0.75}{
\renewcommand{\arraystretch}{1.0}
\begin{tabular}{>{\hspace{1pt}}c<{\hspace{1pt}} >{\hspace{1pt}}c<{\hspace{1pt}} cc cc}
\toprule
\multirow{2}{*}{\makecell[c]{\cellcolor{\hl} Semantic\\\cellcolor{\hl} Attunement\\[-1ex]}} &
\multirow{2}{*}{\makecell[c]{\cellcolor{\hl} Geometric\\\cellcolor{\hl} Consistency\\[-1ex]}}  &
\multicolumn{2}{c}{NTU-60 (48/12)} & \multicolumn{2}{c}{NTU-120 (110/10)}\\
\cmidrule(lr){3-4} \cmidrule(l){5-6} 
 & & ZSL & GZSL & ZSL & GZSL \\
\midrule
\xmark & \xmark & 49.6 & 46.3 & 74.8 & 63.9 \\
\xmark & \cmark & 61.8 & 57.0 & 75.5 & 64.2 \\
\cmark & \xmark & 50.2 & 49.5 & 76.4 & 64.8 \\
\rowcolor{yellow!10} \cmark & \cmark & \textbf{65.3} & \textbf{60.5} & \textbf{79.6} & \textbf{66.1} \\
\bottomrule
\end{tabular}}
\vspace{-5pt}
\end{table}


\vspace{0.3em}
\noindent \textbf{Influence of Components in the Deciding Phase}.
We fix the learning phase and ablate components in the deciding phase to assess their effects. As shown in Table~\ref{table:deciding_component_analysis}, injecting noise into the source notably degrades performance, especially on NTU-60, where clear representations are crucial for discrimination. Conditioning also leads to overfitting on seen domains, reducing generalization to unseen categories. In contrast, contrastive regularization consistently improves performance, though with moderate gains.

\begin{table}[!h]
\caption{Analysis of different components in the deciding phase.}
\label{table:deciding_component_analysis}
\vspace{-5pt}
\centering
\scalebox{0.7}{
\renewcommand{\arraystretch}{1.0}
\begin{tabular}{>{\hspace{1pt}}c<{\hspace{1pt}} >{\hspace{1pt}}c<{\hspace{1pt}} >{\hspace{1pt}}c<{\hspace{1pt}} cc cc}
\toprule
\multirow{2}{*}{\makecell[c]{\cellcolor{\hl} Noise-\\\cellcolor{\hl} Free\\[-1ex]}} &
\multirow{2}{*}{\makecell[c]{\cellcolor{\hl} Condition-\\\cellcolor{\hl} Free\\[-1ex]}} &
\multirow{2}{*}{\makecell[c]{\cellcolor{\hl} Contrastive\\\cellcolor{\hl} Strategy\\[-1ex]}} &
\multicolumn{2}{c}{NTU-60 (48/12)} &
\multicolumn{2}{c}{NTU-120 (110/10)} \\
\cmidrule(lr){4-5} \cmidrule(lr){6-7} 
 & & & ZSL & GZSL & ZSL & GZSL  \\
\midrule
\xmark & \xmark & \xmark & 53.3 & 49.0 & 75.7 & 60.5 \\
\cmark & \xmark & \xmark & 62.2 & 57.9 & 77.1 & 63.4 \\
\xmark & \cmark & \xmark & 55.1 & 51.0 & 77.2 & 63.1 \\
\cmark & \cmark & \xmark & 64.0 & 60.4 & 78.1 & 65.8 \\
\rowcolor{yellow!10} \cmark & \cmark & \cmark & \textbf{65.3} & \textbf{60.5} & \textbf{79.6} & \textbf{66.1} \\
\bottomrule
\end{tabular}}
\vspace{-5pt}
\end{table}


\vspace{0.3em}
\noindent \textbf{Influence of Inference Timestep $t$ Selection}.
As shown in Fig.~\ref{fig:timestep}, the performance remains stable when using smaller timestep values, where category semantics contribute more effectively to the latent embedding $z_t$. However, as the timestep approaches 1, performance gradually degrades because the predicted velocity increasingly depends on the unseen skeleton embedding $z_s$ alone, rather than the semantic prior $z_a$. This reliance amplifies the uncertainty of unseen skeleton samples, leading to a noticeable drop in accuracy.


\vspace{0.3em}
\noindent \textbf{Classifier Comparison}.
As presented in Table~\ref{table:classifiers}, we compare our flow-based classifier with two common alternatives: the linear classifier used in previous generative methods and the similarity-based matching employed in embedding-based approaches. Our classifier consistently outperforms both baselines.

\begin{table}[!h]
\caption{ZSL performance under different classifier types. 
}
\vspace{-5pt}
\label{table:classifiers}
\centering
\scalebox{0.69}{
\renewcommand{\arraystretch}{1.0}
\begin{tabular}{lcccc}
\toprule
\multirow{2}{*}{Types} & \multicolumn{2}{c}{NTU-60 (Xsub)} & \multicolumn{2}{c}{NTU-120 (Xsub)} \\
\cmidrule(r){2-3} \cmidrule(l){4-5}
 & 55/5 Split & 48/12 Split & 110/10 Split & 96/24 Split\\
\midrule
Linear Classifier~\cite{gupta2021syntactically} & 82.7 & \textcolor{blue}{58.0} & 76.5 & 64.4 \\
Similarity Matching~\cite{zhou2023zero} & \textcolor{blue}{83.9} & 56.7 & \textcolor{blue}{77.1} & \textcolor{blue}{64.7} \\
\rowcolor{yellow!10} \textbf{Ours} & \textcolor{red}{\textbf{86.3}} & \textcolor{red}{\textbf{65.3}} & \textcolor{red}{\textbf{79.6}} &  \textcolor{red}{\textbf{66.4}} \\
\bottomrule
\end{tabular}}
\vspace{-5pt}
\end{table}

\subsection{Qualitative Analysis}

\begin{figure}[!h]
\centering
\includegraphics[width=0.99\linewidth]{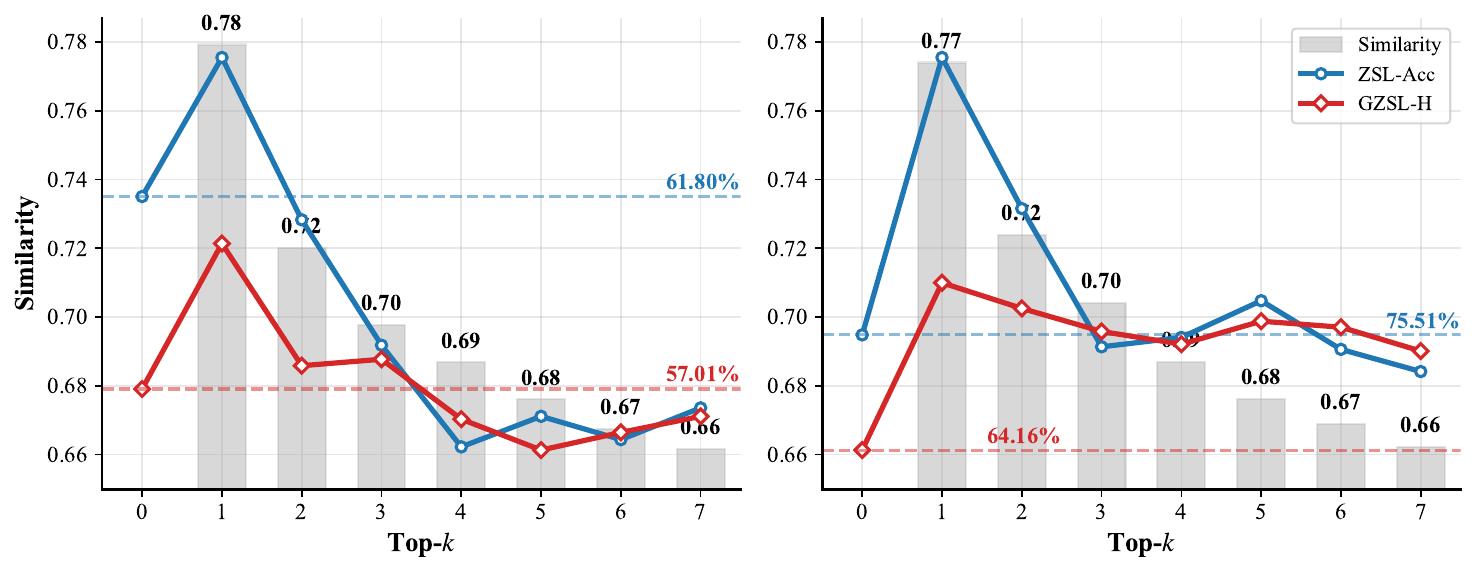} \\
\vspace{-5pt}
\begin{minipage}{0.99\linewidth}
\centering
\begin{tabular}{P{0.5}P{0.4}}
\footnotesize{(a) NTU-60 (48/12)} & 
\footnotesize{(b) NTU-120 (110/10)} \\
\end{tabular}
\end{minipage}
\vspace{-7pt}
\caption{Neighbor selection analysis with corresponding semantic similarity scores on NTU-60 and NTU-120.}
\label{fig:coefficient_k}
\vspace{-10pt}
\end{figure}

\noindent \textbf{Neighbor Selection Analysis}.
As shown in Fig.~\ref{fig:coefficient_k}, introducing the Top-$k$ mechanism notably enhances performance by leveraging local semantic context. However, as $k$ increases, the similarity steadily declines, indicating that distant neighbors are semantically less relevant and less reliable. 
Thus, incorporating too many such neighbors introduces noisy or misleading semantics, which tends to guide the model toward less meaningful regions of the semantic space, rather than reinforcing reasonable alignment.

\begin{figure*}[t!]
\centering
\includegraphics[width=0.99\linewidth]{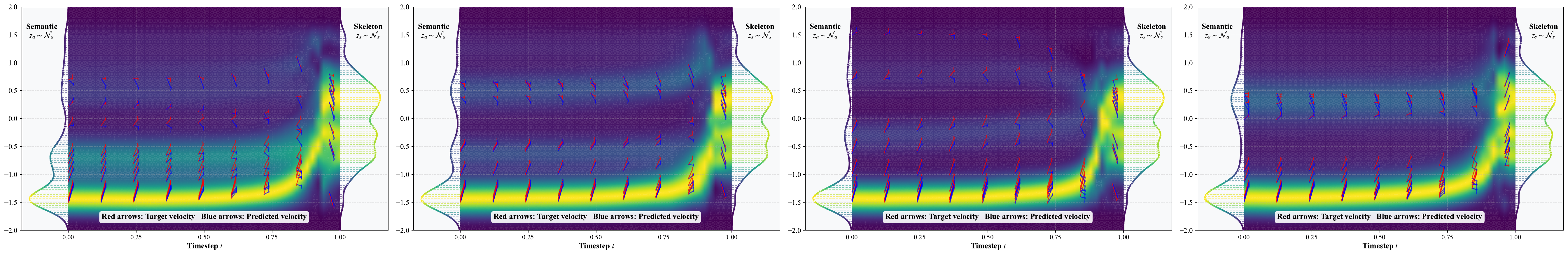} \\
\begin{minipage}{0.99\linewidth}
\centering
\begin{tabular}{P{0.26}P{0.21}P{0.22}P{0.21}}
\footnotesize{\makecell{(a) ``writing'' semantic \\vs. ``put on a hat'' skeleton}} & 
\footnotesize{\makecell{(b) ``put on a hat'' semantic \\vs. ``put on a hat'' skeleton}}  & 
\footnotesize{\makecell{(c) ``jump up'' semantic \\vs. ``put on a hat'' skeleton}} & 
\footnotesize{\makecell{(d) ``touch pocket'' semantic \\vs. ``put on a hat'' skeleton}} \\
\end{tabular}
\end{minipage}
\vspace{-7pt}
\caption{Flow velocity visualization in the deciding phase on NTU-60 (55/5 Split). Each pair shows distribution transport from the semantic (left) to the skeleton (right) space, with \textcolor{red}{red} and \textcolor{blue}{blue} arrows denoting target and predicted velocities (zoom in for a better view).}
\label{fig:flow_distribution_with_velocity}
\vspace{-10pt}
\end{figure*}

\begin{figure*}[t!]
\begin{center}
\begin{tabular}{cc}
    \includegraphics[width=0.6\linewidth,height=0.15\linewidth]{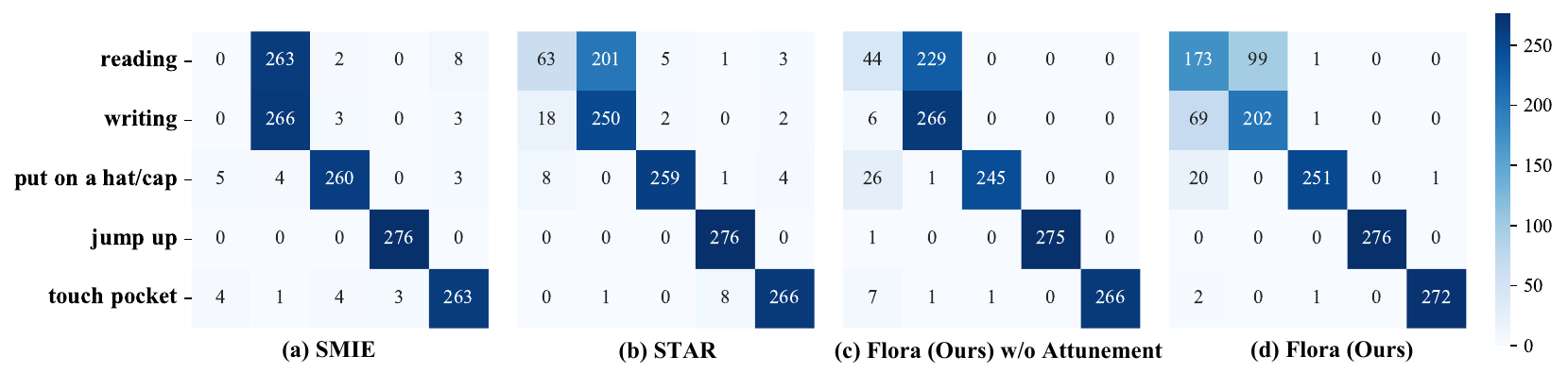} &
    \includegraphics[width=0.35\linewidth,height=0.15\linewidth]{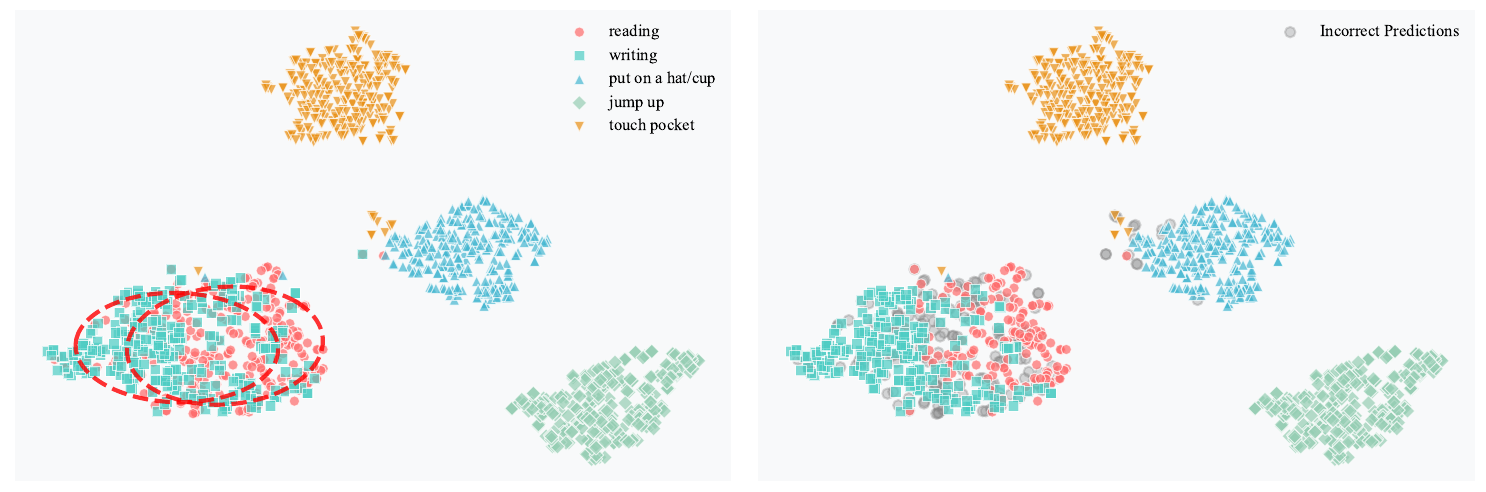} \\
    \vspace{-10pt}
    \scriptsize{(a) Confusion matrix} & 
    \scriptsize{(b) t-SNE visualization} \\
\end{tabular}
\end{center}
\vspace{-10pt}
\caption{Similar action comparison and the corresponding t-SNE visualization (NTU60, 55/5, STAR-based features).}
\label{fig:similar_unseen_comparison}
\end{figure*}

\begin{figure}[h!]
\centering
\includegraphics[width=0.99\linewidth]{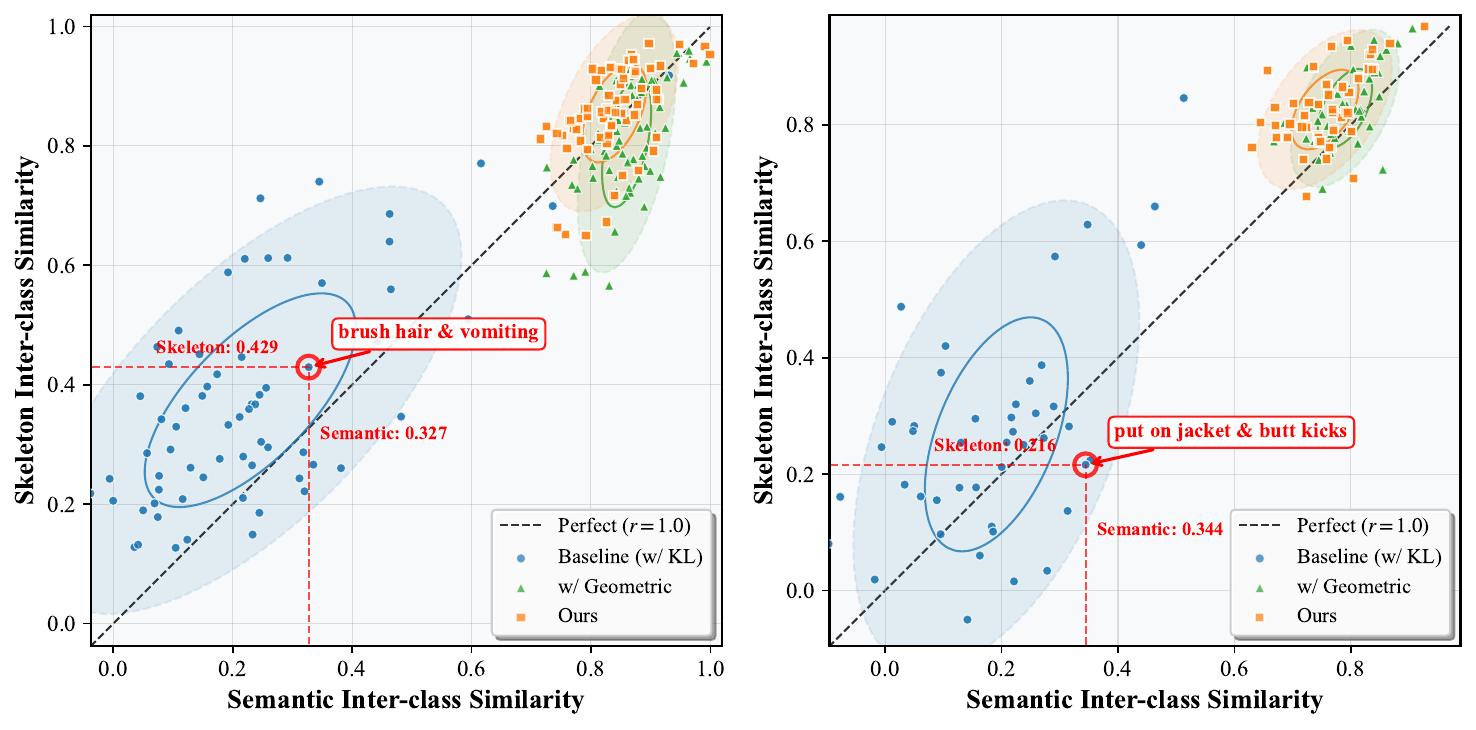} \\
\vspace{-5pt}
\begin{minipage}{0.99\linewidth}
\centering
\begin{tabular}{P{0.5}P{0.4}}
\footnotesize{(a) NTU-60 (48/12)} & 
\footnotesize{(b) NTU-120 (110/10)} \\
\end{tabular}
\end{minipage}
\vspace{-7pt}
\caption{Cross-modal alignment analysis in the learning phase. Each dot represents the inter-class similarity between paired categories in the skeleton and semantic spaces, where proximity to the diagonal indicates stronger structural consistency. \textcolor{blue}{Blue}, \textcolor{green}{green}, and \textcolor{orange}{orange} correspond to the baseline (Sec.~\ref{sec:cross_modal_vae}), geometric consistency, and our full model with semantic attunement, respectively.}
\label{fig:cross_modal_alignment}
\vspace{-10pt}
\end{figure}

\vspace{0.3em}
\noindent \textbf{Cross-modal Alignment Analysis}.
We compute the mean latent embedding of each category in both skeleton and semantic spaces and measure inter-class similarities within each. As shown in Fig.~\ref{fig:cross_modal_alignment}, the baseline (Sec.~\ref{sec:cross_modal_vae}) exhibits poor alignment with scattered points deviating from the diagonal (blue). Replacing the KL divergence with geometric consistency (green) improves structural correspondence, while adding semantic attunement (orange) yields the most coherent and semantically consistent cross-modal alignment.


\vspace{0.3em}
\noindent \textbf{Flow Velocity Analysis}.
We visualize the distribution transport from the semantic source $\mathcal{N}_a$ to the skeleton target $\mathcal{N}_s$ along with the corresponding velocity fields. As shown in Fig.~\ref{fig:flow_distribution_with_velocity}, the transportation paths vary across different semantic–skeleton pairs, forming the foundation for reliable classification. Moreover, the matched semantic–skeleton pairs exhibit the smallest discrepancy between the predicted and ground-truth velocities (Fig.~\ref{fig:flow_distribution_with_velocity}(b)). Additionally, the prediction error increases as the timestep approaches 1, which is consistent with the observation in Fig.~\ref{fig:timestep}.



\subsection{Discussions}

In Fig.~\ref{fig:similar_unseen_comparison}, our method still struggles to distinguish highly similar unseen categories such as ``reading'' and ``writing'' completely. Although improvement by semantic attunement, the overlapped skeleton features still exist. Since these actions share nearly identical motion patterns, separating them based solely on the seen skeletons and their associated semantics remains difficult. A promising direction is to develop skeleton-specific semantics that more precisely capture subtle motion cues, rather than relying on current action-level semantics, which often introduce ambiguity. Additional discussions are provided in the \textcolor{magenta}{Appendix~\ref{app:additional_discussions}}.

\section{Conclusion}
\label{sec:conclusion}
In this paper, we present \texttt{\textbf{Flora}}, a novel framework designed to overcome the key limitations of the conventional ``align-then-classify'' paradigm. By integrating adjacent inter-class contextual semantics with a geometric consistency objective, \texttt{\textbf{Flora}} achieves stable and direction-aware point-to-region alignment. Moreover, the proposed distribution-aware flow classifier enables fine-grained recognition with plug-and-play flexibility, supporting noise-free, condition-free, and boundary-free decision-making. These advancements substantially enhance generalizability, even with simple architectures and limited training data, showcasing strong potential for further zero-shot skeleton action recognition research.

\clearpage
\section*{Acknowledgments}
This research was supported by the Hong Kong RGC General Research Fund (Grant Nos. 15221123, 15216424, and 15211525) and the Hong Kong PolyU Internal Research Fund (Grant Nos. P0058468 and P0056171).
{
    \small
    \bibliographystyle{ieeenat_fullname}
    \bibliography{main}

\begin{thebibliography}{49}
\providecommand{\natexlab}[1]{#1}
\providecommand{\url}[1]{\texttt{#1}}
\expandafter\ifx\csname urlstyle\endcsname\relax
  \providecommand{\doi}[1]{doi: #1}\else
  \providecommand{\doi}{doi: \begingroup \urlstyle{rm}\Url}\fi

\bibitem[Albergo and Vanden-Eijnden(2022)]{albergo2022building}
Michael~S Albergo and Eric Vanden-Eijnden.
\newblock Building normalizing flows with stochastic interpolants.
\newblock \emph{arXiv preprint arXiv:2209.15571}, 2022.

\bibitem[Chen et~al.(2023)Chen, Yang, Wang, Wang, Cheng, and Wang]{chen2023stformer}
Yang Chen, Shuang Yang, Yingying Wang, Guorong Wang, Hong Cheng, and Ling Wang.
\newblock Stformer: Spatial-temporal transformer for early warning of unplanned extubation in icu.
\newblock In \emph{2023 45th Annual International Conference of the IEEE Engineering in Medicine \& Biology Society (EMBC)}, pages 1--4. IEEE, 2023.

\bibitem[Chen et~al.(2024)Chen, Guo, He, Lu, and Wang]{chen2024fine}
Yang Chen, Jingcai Guo, Tian He, Xiaocheng Lu, and Ling Wang.
\newblock Fine-grained side information guided dual-prompts for zero-shot skeleton action recognition.
\newblock In \emph{Proceedings of the 32nd ACM International Conference on Multimedia}, pages 778--786, 2024.

\bibitem[Chen et~al.(2025)Chen, Guo, Guo, and Tao]{chen2025neuron}
Yang Chen, Jingcai Guo, Song Guo, and Dacheng Tao.
\newblock Neuron: Learning context-aware evolving representations for zero-shot skeleton action recognition.
\newblock In \emph{Proceedings of the Computer Vision and Pattern Recognition Conference}, pages 8721--8730, 2025.

\bibitem[Chen et~al.(2026)Chen, Guo, Li, Rao, and Guo]{chen2026star++}
Yang Chen, Jingcai Guo, Miaoge Li, Zhijie Rao, and Song Guo.
\newblock {STAR++: Region-aware Conditional Semantics via Interpretable Side Information for Zero-Shot Skeleton Action Recognition}.
\newblock \emph{IEEE Transactions on Circuits and Systems for Video Technology}, 2026.

\bibitem[Cheng et~al.(2020)Cheng, Zhang, He, Chen, Cheng, and Lu]{cheng2020skeleton}
Ke Cheng, Yifan Zhang, Xiangyu He, Weihan Chen, Jian Cheng, and Hanqing Lu.
\newblock Skeleton-based action recognition with shift graph convolutional network.
\newblock In \emph{Proceedings of the IEEE/CVF conference on computer vision and pattern recognition}, pages 183--192, 2020.

\bibitem[Chunhui et~al.(2017)Chunhui, Yueyu, Yanghao, Sijie, and Jiaying]{liu2017pku}
Liu Chunhui, Hu Yueyu, Li Yanghao, Song Sijie, and Liu Jiaying.
\newblock Pku-mmd: A large scale benchmark for continuous multi-modal human action understanding.
\newblock \emph{arXiv preprint arXiv:1703.07475}, 2017.

\bibitem[Do and Kim(2024)]{do2024tdsm}
Jeonghyeok Do and Munchurl Kim.
\newblock Tdsm: Triplet diffusion for skeleton-text matching in zero-shot action recognition.
\newblock \emph{arXiv preprint arXiv:2411.10745}, 2024.

\bibitem[Esser et~al.(2024)Esser, Kulal, Blattmann, Entezari, M{\"u}ller, Saini, Levi, Lorenz, Sauer, Boesel, et~al.]{esser2024scaling}
Patrick Esser, Sumith Kulal, Andreas Blattmann, Rahim Entezari, Jonas M{\"u}ller, Harry Saini, Yam Levi, Dominik Lorenz, Axel Sauer, Frederic Boesel, et~al.
\newblock Scaling rectified flow transformers for high-resolution image synthesis.
\newblock In \emph{Forty-first international conference on machine learning}, 2024.

\bibitem[Gao et~al.(2025)Gao, Zhao, Lee, Chuang, Zhou, Wang, Zhao, Zhang, and Soltani]{gao2025vita}
Dechen Gao, Boqi Zhao, Andrew Lee, Ian Chuang, Hanchu Zhou, Hang Wang, Zhe Zhao, Junshan Zhang, and Iman Soltani.
\newblock Vita: Vision-to-action flow matching policy.
\newblock \emph{arXiv preprint arXiv:2507.13231}, 2025.

\bibitem[Gupta et~al.(2021)Gupta, Sharma, and Sarvadevabhatla]{gupta2021syntactically}
Pranay Gupta, Divyanshu Sharma, and Ravi~Kiran Sarvadevabhatla.
\newblock Syntactically guided generative embeddings for zero-shot skeleton action recognition.
\newblock In \emph{2021 IEEE International Conference on Image Processing (ICIP)}, pages 439--443. IEEE, 2021.

\bibitem[He et~al.(2025)He, Yu, Liu, and Chen]{he2025flowtok}
Ju He, Qihang Yu, Qihao Liu, and Liang-Chieh Chen.
\newblock Flowtok: Flowing seamlessly across text and image tokens.
\newblock \emph{arXiv preprint arXiv:2503.10772}, 2025.

\bibitem[He et~al.(2024)He, Chen, Wang, and Cheng]{he2024expert}
Tian He, Yang Chen, Ling Wang, and Hong Cheng.
\newblock An expert-knowledge-based graph convolutional network for skeleton-based physical rehabilitation exercises assessment.
\newblock \emph{IEEE Transactions on Neural Systems and Rehabilitation Engineering}, 32:\penalty0 1916--1925, 2024.

\bibitem[Higgins et~al.(2017)Higgins, Matthey, Pal, Burgess, Glorot, Botvinick, Mohamed, and Lerchner]{higgins2017beta}
Irina Higgins, Loic Matthey, Arka Pal, Christopher Burgess, Xavier Glorot, Matthew Botvinick, Shakir Mohamed, and Alexander Lerchner.
\newblock beta-vae: Learning basic visual concepts with a constrained variational framework.
\newblock In \emph{International conference on learning representations}, 2017.

\bibitem[Hong et~al.(2021)Hong, Fisher, Gharbi, and Fatahalian]{hong2021video}
James Hong, Matthew Fisher, Micha{\"e}l Gharbi, and Kayvon Fatahalian.
\newblock Video pose distillation for few-shot, fine-grained sports action recognition.
\newblock In \emph{Proceedings of the IEEE/CVF International Conference on Computer Vision}, pages 9254--9263, 2021.

\bibitem[Hu et~al.(2024)Hu, Wu, Asano, Mettes, Fernando, Ommer, and Snoek]{hu2024flow}
Vincent Hu, Di Wu, Yuki Asano, Pascal Mettes, Basura Fernando, Bj{\"o}rn Ommer, and Cees Snoek.
\newblock Flow matching for conditional text generation in a few sampling steps.
\newblock In \emph{Proceedings of the 18th Conference of the European Chapter of the Association for Computational Linguistics (Volume 2: Short Papers)}, pages 380--392, 2024.

\bibitem[Hu et~al.(2023)Hu, Yin, Ma, Chen, Fernando, Asano, Gavves, Mettes, Ommer, and Snoek]{hu2023motion}
Vincent~Tao Hu, Wenzhe Yin, Pingchuan Ma, Yunlu Chen, Basura Fernando, Yuki~M Asano, Efstratios Gavves, Pascal Mettes, Bjorn Ommer, and Cees~GM Snoek.
\newblock Motion flow matching for human motion synthesis and editing.
\newblock \emph{arXiv preprint arXiv:2312.08895}, 2023.

\bibitem[Hubert~Tsai et~al.(2017)Hubert~Tsai, Huang, and Salakhutdinov]{hubert2017learning}
Yao-Hung Hubert~Tsai, Liang-Kang Huang, and Ruslan Salakhutdinov.
\newblock Learning robust visual-semantic embeddings.
\newblock In \emph{Proceedings of the IEEE International conference on Computer Vision}, pages 3571--3580, 2017.

\bibitem[Jasani and Mazagonwalla(2019)]{jasani2019skeleton}
Bhavan Jasani and Afshaan Mazagonwalla.
\newblock Skeleton based zero shot action recognition in joint pose-language semantic space.
\newblock \emph{arXiv preprint arXiv:1911.11344}, 2019.

\bibitem[Kingma and Welling(2013)]{kingma2013auto}
Diederik~P Kingma and Max Welling.
\newblock Auto-encoding variational bayes.
\newblock \emph{arXiv preprint arXiv:1312.6114}, 2013.

\bibitem[Kuang et~al.(2025)Kuang, Wang, Han, Zhang, and Gui]{kuang2025zero}
Jidong Kuang, Hongsong Wang, Chaolei Han, Yang Zhang, and Jie Gui.
\newblock Zero-shot skeleton-based action recognition with dual visual-text alignment.
\newblock \emph{Pattern Recognition}, page 112342, 2025.

\bibitem[Li et~al.(2023)Li, Jia, Zhang, Ma, and Wang]{li2023multi}
Ming-Zhe Li, Zhen Jia, Zhang Zhang, Zhanyu Ma, and Liang Wang.
\newblock Multi-semantic fusion model for generalized zero-shot skeleton-based action recognition.
\newblock In \emph{International Conference on Image and Graphics}, pages 68--80. Springer, 2023.

\bibitem[Li et~al.(2024)Li, Wei, Chen, Yu, Yang, and Hsu]{li2024sa}
Sheng-Wei Li, Zi-Xiang Wei, Wei-Jie Chen, Yi-Hsin Yu, Chih-Yuan Yang, and Jane Yung-jen Hsu.
\newblock Sa-dvae: Improving zero-shot skeleton-based action recognition by disentangled variational autoencoders.
\newblock In \emph{European Conference on Computer Vision}, pages 447--462. Springer, 2024.

\bibitem[Lipman et~al.(2022)Lipman, Chen, Ben-Hamu, Nickel, and Le]{lipman2022flow}
Yaron Lipman, Ricky~TQ Chen, Heli Ben-Hamu, Maximilian Nickel, and Matt Le.
\newblock Flow matching for generative modeling.
\newblock \emph{arXiv preprint arXiv:2210.02747}, 2022.

\bibitem[Lipman et~al.(2024)Lipman, Havasi, Holderrieth, Shaul, Le, Karrer, Chen, Lopez-Paz, Ben-Hamu, and Gat]{lipman2024flow}
Yaron Lipman, Marton Havasi, Peter Holderrieth, Neta Shaul, Matt Le, Brian Karrer, Ricky~TQ Chen, David Lopez-Paz, Heli Ben-Hamu, and Itai Gat.
\newblock Flow matching guide and code.
\newblock \emph{arXiv preprint arXiv:2412.06264}, 2024.

\bibitem[Liu et~al.(2025{\natexlab{a}})Liu, Niu, Zeng, Liu, Hu, and Song]{liu2025beyond}
Hongjie Liu, Yingchun Niu, Kun Zeng, Chun Liu, Mengjie Hu, and Qing Song.
\newblock Beyond-skeleton: Zero-shot skeleton action recognition enhanced by supplementary rgb visual information.
\newblock \emph{Expert Systems with Applications}, 273:\penalty0 126814, 2025{\natexlab{a}}.

\bibitem[Liu et~al.(2019)Liu, Shahroudy, Perez, Wang, Duan, and Kot]{liu2019ntu}
Jun Liu, Amir Shahroudy, Mauricio Perez, Gang Wang, Ling-Yu Duan, and Alex~C Kot.
\newblock Ntu rgb+ d 120: A large-scale benchmark for 3d human activity understanding.
\newblock \emph{IEEE transactions on pattern analysis and machine intelligence}, 42\penalty0 (10):\penalty0 2684--2701, 2019.

\bibitem[Liu et~al.(2025{\natexlab{b}})Liu, Yin, Yuille, Brown, and Singh]{liu2025flowing}
Qihao Liu, Xi Yin, Alan Yuille, Andrew Brown, and Mannat Singh.
\newblock Flowing from words to pixels: A noise-free framework for cross-modality evolution.
\newblock In \emph{Proceedings of the Computer Vision and Pattern Recognition Conference}, pages 2755--2765, 2025{\natexlab{b}}.

\bibitem[Liu et~al.(2022)Liu, Gong, and Liu]{liu2022flow}
Xingchao Liu, Chengyue Gong, and Qiang Liu.
\newblock Flow straight and fast: Learning to generate and transfer data with rectified flow.
\newblock \emph{arXiv preprint arXiv:2209.03003}, 2022.

\bibitem[Ma et~al.(2024)Ma, Song, Zhuang, Hao, and King]{ma2024survey}
Yueen Ma, Zixing Song, Yuzheng Zhuang, Jianye Hao, and Irwin King.
\newblock A survey on vision-language-action models for embodied ai.
\newblock \emph{arXiv preprint arXiv:2405.14093}, 2024.

\bibitem[Mishra et~al.(2024)Mishra, Mihailidis, and Khan]{mishra2024skeletal}
Pratik~K Mishra, Alex Mihailidis, and Shehroz~S Khan.
\newblock Skeletal video anomaly detection using deep learning: Survey, challenges, and future directions.
\newblock \emph{IEEE Transactions on Emerging Topics in Computational Intelligence}, 8\penalty0 (2):\penalty0 1073--1085, 2024.

\bibitem[Peebles and Xie(2023)]{peebles2023scalable}
William Peebles and Saining Xie.
\newblock Scalable diffusion models with transformers.
\newblock In \emph{Proceedings of the IEEE/CVF international conference on computer vision}, pages 4195--4205, 2023.

\bibitem[Polyak et~al.(2024)Polyak, Zohar, Brown, Tjandra, Sinha, Lee, Vyas, Shi, Ma, Chuang, et~al.]{polyak2024movie}
Adam Polyak, Amit Zohar, Andrew Brown, Andros Tjandra, Animesh Sinha, Ann Lee, Apoorv Vyas, Bowen Shi, Chih-Yao Ma, Ching-Yao Chuang, et~al.
\newblock Movie gen: A cast of media foundation models.
\newblock \emph{arXiv preprint arXiv:2410.13720}, 2024.

\bibitem[Radford et~al.(2021)Radford, Kim, Hallacy, Ramesh, Goh, Agarwal, Sastry, Askell, Mishkin, Clark, et~al.]{radford2021learning}
Alec Radford, Jong~Wook Kim, Chris Hallacy, Aditya Ramesh, Gabriel Goh, Sandhini Agarwal, Girish Sastry, Amanda Askell, Pamela Mishkin, Jack Clark, et~al.
\newblock Learning transferable visual models from natural language supervision.
\newblock In \emph{International conference on machine learning}, pages 8748--8763. PmLR, 2021.

\bibitem[Rao et~al.(2025)Rao, Wu, Jiang, Zhang, Wang, and Xie]{rao2025towards}
Jiayuan Rao, Haoning Wu, Hao Jiang, Ya Zhang, Yanfeng Wang, and Weidi Xie.
\newblock Towards universal soccer video understanding.
\newblock In \emph{Proceedings of the Computer Vision and Pattern Recognition Conference}, pages 8384--8394, 2025.

\bibitem[Sato et~al.(2023)Sato, Hachiuma, and Sekii]{sato2023prompt}
Fumiaki Sato, Ryo Hachiuma, and Taiki Sekii.
\newblock Prompt-guided zero-shot anomaly action recognition using pretrained deep skeleton features.
\newblock In \emph{Proceedings of the IEEE/CVF conference on computer vision and pattern recognition}, pages 6471--6480, 2023.

\bibitem[Schonfeld et~al.(2019)Schonfeld, Ebrahimi, Sinha, Darrell, and Akata]{schonfeld2019generalized}
Edgar Schonfeld, Sayna Ebrahimi, Samarth Sinha, Trevor Darrell, and Zeynep Akata.
\newblock Generalized zero-shot learning via aligned variational autoencoders.
\newblock In \emph{Proceedings of the IEEE/CVF Conference on Computer Vision and Pattern Recognition Workshops}, pages 54--57, 2019.

\bibitem[Shahroudy et~al.(2016)Shahroudy, Liu, Ng, and Wang]{shahroudy2016ntu}
Amir Shahroudy, Jun Liu, Tian-Tsong Ng, and Gang Wang.
\newblock Ntu rgb+ d: A large scale dataset for 3d human activity analysis.
\newblock In \emph{Proceedings of the IEEE conference on computer vision and pattern recognition}, pages 1010--1019, 2016.

\bibitem[Stoica et~al.(2025)Stoica, Ramanujan, Fan, Farhadi, Krishna, and Hoffman]{stoica2025contrastive}
George Stoica, Vivek Ramanujan, Xiang Fan, Ali Farhadi, Ranjay Krishna, and Judy Hoffman.
\newblock Contrastive flow matching.
\newblock \emph{arXiv preprint arXiv:2506.05350}, 2025.

\bibitem[Vyas et~al.(2023)Vyas, Shi, Le, Tjandra, Wu, Guo, Zhang, Zhang, Adkins, Ngan, et~al.]{vyas2023audiobox}
Apoorv Vyas, Bowen Shi, Matthew Le, Andros Tjandra, Yi-Chiao Wu, Baishan Guo, Jiemin Zhang, Xinyue Zhang, Robert Adkins, William Ngan, et~al.
\newblock Audiobox: Unified audio generation with natural language prompts.
\newblock \emph{arXiv preprint arXiv:2312.15821}, 2023.

\bibitem[Wray et~al.(2019)Wray, Larlus, Csurka, and Damen]{wray2019fine}
Michael Wray, Diane Larlus, Gabriela Csurka, and Dima Damen.
\newblock Fine-grained action retrieval through multiple parts-of-speech embeddings.
\newblock In \emph{Proceedings of the IEEE/CVF international conference on computer vision}, pages 450--459, 2019.

\bibitem[Wu et~al.(2025)Wu, Guo, Chen, Xue, and Lu]{wu2025frequency}
Wenhan Wu, Zhishuai Guo, Chen Chen, Hongfei Xue, and Aidong Lu.
\newblock Frequency-semantic enhanced variational autoencoder for zero-shot skeleton-based action recognition.
\newblock \emph{arXiv preprint arXiv:2506.22179}, 2025.

\bibitem[Xu et~al.(2025)Xu, Gao, Li, and Gao]{xu2025information}
Haojun Xu, Yan Gao, Jie Li, and Xinbo Gao.
\newblock An information compensation framework for zero-shot skeleton-based action recognition.
\newblock \emph{IEEE Transactions on Multimedia}, 2025.

\bibitem[Zhang et~al.(2024)Zhang, Sui, and Yeung-Levy]{zhang2024connect}
Yuhui Zhang, Elaine Sui, and Serena Yeung-Levy.
\newblock Connect, collapse, corrupt: Learning cross-modal tasks with uni-modal data.
\newblock \emph{arXiv preprint arXiv:2401.08567}, 2024.

\bibitem[Zhou et~al.(2025)Zhou, Zhang, You, Hu, Tan, and Liu]{zhou2025zero}
Kai Zhou, Shuhai Zhang, Zeng You, Jinwu Hu, Mingkui Tan, and Fei Liu.
\newblock Zero-shot skeleton-based action recognition with prototype-guided feature alignment.
\newblock \emph{arXiv preprint arXiv:2507.00566}, 2025.

\bibitem[Zhou et~al.(2023)Zhou, Qiang, Rao, Lin, Su, and Wang]{zhou2023zero}
Yujie Zhou, Wenwen Qiang, Anyi Rao, Ning Lin, Bing Su, and Jiaqi Wang.
\newblock Zero-shot skeleton-based action recognition via mutual information estimation and maximization.
\newblock In \emph{Proceedings of the 31st ACM international conference on multimedia}, pages 5302--5310, 2023.

\bibitem[Zhu et~al.(2024)Zhu, Ke, Gong, and Bailey]{zhu2024part}
Anqi Zhu, Qiuhong Ke, Mingming Gong, and James Bailey.
\newblock Part-aware unified representation of language and skeleton for zero-shot action recognition.
\newblock In \emph{Proceedings of the IEEE/CVF Conference on Computer Vision and Pattern Recognition}, pages 18761--18770, 2024.

\bibitem[Zhu et~al.(2025{\natexlab{a}})Zhu, Zhu, Bailey, Gong, and Ke]{zhu2025semantic}
Anqi Zhu, Jingmin Zhu, James Bailey, Mingming Gong, and Qiuhong Ke.
\newblock Semantic-guided cross-modal prompt learning for skeleton-based zero-shot action recognition.
\newblock In \emph{Proceedings of the Computer Vision and Pattern Recognition Conference}, pages 13876--13885, 2025{\natexlab{a}}.

\bibitem[Zhu et~al.(2025{\natexlab{b}})Zhu, Shu, Huang, and Tang]{zhu2025prompt}
Xingyu Zhu, Xiangbo Shu, Peng Huang, and Jinhui Tang.
\newblock Prompt-guided prototype-aware commonality and discrimination learning for zero-shot skeleton-based action recognition.
\newblock \emph{IEEE Transactions on Multimedia}, 2025{\natexlab{b}}.

\end{thebibliography}
}

\clearpage
\setcounter{page}{1}
\maketitlesupplementary
\appendix

\section*{\underline{Appendix Roadmap}}
The supplementary material is organized into the following sections:
\begin{itemize}
    \item Sec.~\ref{app:datasets}: \textbf{Datasets.}
        \begin{itemize}
            \item[-] \myref{app:ntu60}{NTU RGB+D 60.}
            \item[-] \myref{app:ntu120}{NTU RGB+D 120.}
            \item[-] \myref{app:pku51}{PKU-MMD.}
        \end{itemize} 
    \item Sec.~\ref{app:dataset_splits}: \textbf{Dataset Seen-Unseen Split Details.}
        \begin{itemize}
            \item[-] \myref{app:basic_split}{Basic Seen-Unseen Split Details.}
            \item[-] \myref{app:challenging_split}{Challenging Seen-Unseen Split Details.}
            \item[-] \myref{app:random_split}{Random Seen-Unseen Split Details.}
        \end{itemize} 
    \item Sec.~\ref{app:implementation_details}: \textbf{Implementation Details.}
    \item Sec.~\ref{app:addtional_sota}: \textbf{Additional Performance Comparison}
        \begin{itemize}
            \item[-] \myref{app:sota_basic_ii}{Basic Split Benchmark Evaluation II.}
            \item[-] \myref{app:sota_random_ii}{Random Split Benchmark Evaluation II.}
            \item[-] \myref{app:sota_challenge}{More Challenging Seen-Unseen Evaluation.}
            \item[-] \myref{app:inference_time}{Per-instance Inference Time Comparison}
        \end{itemize} 
    \item Sec.~\ref{app:additional_ablation_studies}: \textbf{Additional Ablation Studies}
        \begin{itemize}
            \item[-] \myref{app:ab_learning_and_deciding_phases}{Influence of Learning and Deciding Phases.}
            \item[-] \myref{app:ab_text}{Influence of Text Encoders.}
            \item[-] \myref{app:ab_token}{Influence of Token Numbers $M_a$.}
            \item[-] \myref{app:ab_coefficient_tau}{Influence of coefficient $\tau$.}
            \item[-] \myref{app:ab_threshold_gamma}{Influence of Threshold $\gamma$ in GZSL Prediction.}
            \item[-] \myref{app:ab_coefficient_align}{Influence of Distribution Alignment Coefficient $\lambda_\mathrm{Align}$ in the Learning Phase.}
            \item[-] \myref{app:ab_coefficient_contrastive_fm}{Influence of Contrastive Regularization Coefficient $\lambda_\mathrm{Flow}$ in the Deciding Phase.}
            \item[-] \myref{app:ab_timestep_sampling_types}{Influence of Timestep Sampling Types in the Deciding Phase.}
            \item[-] \myref{app:fm_backbone}{Influence of Flow Matching Backbone.}
            \item[-] \myref{app:ab_flow_directions}{Influence of Flow Directions in the Deciding Phase.}
        \end{itemize} 
    \item Sec.~\ref{app:additional_discussions}: \textbf{Additional Discussions}
        \begin{itemize}
            \item[-] \myref{app:ad_skeleton}{Skeleton Perspective.}
            \item[-] \myref{app:ad_semantic}{Semantic Perspective.}
            \item[-] \myref{app:ad_algorithm}{Algorithm Perspective.}
        \end{itemize} 
\end{itemize}

\section{Datasets}
\label{app:datasets}

\noindent\textbf{NTU RGB+D 60}~\cite{shahroudy2016ntu}.
\label{app:ntu60}
The dataset consists of 56,880 skeleton sequences spanning 60 action categories, performed by 40 subjects and captured from three distinct camera views. It has two standard evaluation protocols, including cross-subject (Xsub) and cross-view (Xview). (i) In the Xsub setting, all sequences are split according to subject identities, with 20 subjects used for training and the remaining 20 for testing. (ii) In the Xview setting, the data are divided by camera viewpoints, where view2 and view3 are used for training, and view1 is used for testing.

\vspace{0.3em}
\noindent\textbf{NTU RGB+D 120}~\cite{liu2019ntu}.
\label{app:ntu120}
The dataset is an extended version of the NTU RGB+D 60~\cite{shahroudy2016ntu} dataset. Compared with the former, this dataset comprises 114,480 sequences covering 120 action categories. Meanwhile, it also provides two official evaluation protocols, including the cross-subject (Xsub) and cross-setup (Xset). (i) In the Xsub setting, sequences from 53 subjects are used for training, while those from the remaining subjects are reserved for testing. (ii) In the Xset setting, data captured using cameras with even IDs are used for training, and those with odd IDs are used for testing.

\vspace{0.3em}
\noindent\textbf{PKU-MMD}~\cite{liu2017pku}.
\label{app:pku51}
The dataset contains approximately 20,000 skeleton sequences across 51 action categories and is organized into two phases with progressively increasing difficulty. Specifically, it also provides two official evaluation protocols, including the cross-subject (Xsub) and cross-view (Xview). (i) In the Xsub setting, sequences from 57 subjects are used for training, while those from the remaining 9 subjects are reserved for testing. (ii) In the Xview setting, data captured from the middle and right camera views are used for training, and those from the left view are used for testing. Following~\cite{zhou2023zero, chen2024fine, chen2025neuron}, we conduct all experiments on the first phase of the dataset.


\section{Dataset Seen-Unseen Split Details}
\label{app:dataset_splits}


\noindent\textbf{Basic Seen-Unseen Split Details}.
\label{app:basic_split}
Table~\ref{table:basic_splits} summarizes the basic seen–unseen splits used in our experiments. The 55/5 and 48/12 splits for NTU-60, as well as the 110/10 and 96/24 splits for NTU-120, follow the official settings in~\cite{gupta2021syntactically}. For PKU-MMD, we follow~\cite{chen2024fine} using the 46/5 and 39/12 splits.

\begin{table}[!h]
\centering
\caption{Basic seen-unseen split details.}
\label{table:basic_splits}
\scalebox{0.82}{
\renewcommand{\arraystretch}{1.0}
\begin{tabular}{ll}
\toprule
Dataset & Split Details (Unseen Category Indices) \\
\midrule
\rowcolor{gray!10} \multicolumn{2}{l}{\textit{NTU-60}~\cite{shahroudy2016ntu}:} \\
55/5 Split~\cite{gupta2021syntactically} & [10, 11, 19, 26, 56] \\
48/12 Split~\cite{gupta2021syntactically} & [3, 5, 9, 12, 15, 40, 42, 47, 51, 56, 58, 59] \\ 
\midrule
\rowcolor{gray!10} \multicolumn{2}{l}{\textit{NTU-120}~\cite{liu2019ntu}:} \\
110/10 Split~\cite{gupta2021syntactically} & [4, 13, 37, 43, 49, 65, 88, 95, 99, 106] \\
\multirow{2}{*}{96/24 Split~\cite{gupta2021syntactically}} & [5, 9, 11, 16, 18, 20, 22, 29, 35, 39, 45, 49, 59,   \\
 & 68, 70, 81, 84, 87, 93, 94, 104, 113, 114, 119] \\
\midrule
\rowcolor{gray!10} \multicolumn{2}{l}{\textit{PKU-MMD}~\cite{liu2017pku}:} \\
46/5 Split~\cite{chen2024fine} & [1, 9, 20, 34, 50] \\
39/12 Split~\cite{chen2024fine} & [3, 7, 11, 15, 19, 21, 25, 31, 33, 36, 43, 48]\\
\bottomrule
\end{tabular}}
\end{table}

\begin{table*}[!h]
\centering
\caption{Challenging seen-unseen split details.}
\label{table:challenging_splits}
\scalebox{0.9}{
\renewcommand{\arraystretch}{1.0}
\begin{tabular}{ll}
\toprule
Dataset & Split Details (Unseen Category Indices) \\
\midrule
\rowcolor{gray!10} \multicolumn{2}{l}{\textit{NTU-60}~\cite{shahroudy2016ntu}:} \\
40/20 Split~\cite{zhu2024part} & [0, 12, 13, 14, 15, 16, 17, 22, 23, 26, 29, 30, 31, 35, 36, 42, 43, 48, 56, 57] \\
30/30 Split~\cite{zhu2024part} & [0, 1, 2, 6, 7, 8, 10, 12, 13, 15, 16, 18, 20, 21, 25, 26, 27, 31, 32, 33, 39, 42, 45, 47, 48, 51, 52, 55, 58, 59] \\
\midrule
\rowcolor{gray!10} \multicolumn{2}{l}{\textit{NTU-120}~\cite{liu2019ntu}:} \\
\multirow{2}{*}{80/40 Split~\cite{zhu2024part}} & [11, 12, 18, 22, 23, 26, 28, 34, 37, 38, 42, 44, 46, 47, 48, 57, 59, 64, 66, 70, 73, 74, 75, 83, 86, 90, 92, 93, 95, 96, \\
 & 102, 104, 107, 108, 110, 112, 115, 116, 118, 119] \\
\multirow{2}{*}{60/60 Split~\cite{zhu2024part}} & [0, 1, 4, 6, 7, 8, 9, 17, 18, 21, 23, 25, 26, 28, 30, 32, 33, 34, 37, 38, 39, 40, 41, 42, 44, 45, 50, 51, 52, 53, 56, 61, \\
 & 62, 65, 67, 68, 69, 70, 74, 77, 78, 81, 83, 87, 89, 90, 91, 92, 94, 95, 96, 97, 100, 101, 109, 111, 114, 115, 116, 118]\\
\bottomrule
\end{tabular}}
\end{table*}


\noindent\textbf{Challenging Seen-Unseen Split Details}.
\label{app:challenging_split}
Table~\ref{table:challenging_splits} summarizes the challenging seen–unseen splits used in our experiments. The 40/20 and 30/30 splits for NTU-60 and the 80/40 and 60/60 splits for NTU-120 are adopted from~\cite{zhu2024part}.


\vspace{0.3em}
\noindent\textbf{Random Seen-Unseen Split Details}.
\label{app:random_split}
Table~\ref{table:random_splits} summarizes the three random seen–unseen splits proposed in SA-DAVE~\cite{li2024sa} and STAR-SMIE~\cite{chen2024fine,zhou2023zero}. The SA-DAVE benchmark is evaluated using ST-GCN features, whereas the STAR-SMIE benchmark employs Shift-GCN features. Notably, STAR-SMIE combines the STAR~\cite{chen2024fine} and SMIE~\cite{zhou2023zero} settings, where STAR defines the PKU-MMD I random splits and SMIE defines the NTU-60 and NTU-120 random splits.

\begin{table}[!h]
\centering
\caption{Three random seen–unseen splits proposed by SA-DAVE~\cite{li2024sa} and STAR-SMIE~\cite{chen2024fine,zhou2023zero}.}
\label{table:random_splits}
\scalebox{0.85}{
\renewcommand{\arraystretch}{1.0}
\begin{tabular}{ll}
\toprule
Dataset & Split Details (Unseen Category Indices) \\
\midrule
\rowcolor{gray!10} \multicolumn{2}{l}{\textit{SA-DAVE}~\cite{li2024sa}:} \\
\multirow{3}{*}{\makecell[l]{NTU-60~\cite{shahroudy2016ntu} \\ (55/5 Split)}} & \ding{202}: [0, 8, 15, 28, 46] \\
 & \ding{203}: [15, 19, 23, 47, 50] \\ 
 & \ding{204}: [29, 37, 38, 45, 55] \\
\hdashline
\multirow{3}{*}{\makecell[l]{NTU-120~\cite{liu2019ntu}\\ (110/10 Split)}} &  \ding{202}: [0, 4, 6, 7, 24, 37, 54, 59, 97, 113] \\
 & \ding{203}: [63, 79, 86, 92, 98, 100, 103, 110, 111, 117] \\
 & \ding{204}: [9, 14, 17, 44, 60, 75, 81, 89, 108, 110] \\
\hdashline
\multirow{3}{*}{\makecell[l]{PKU-MMD I~\cite{liu2017pku}\\ (46/5 Split)}} & \ding{202}: [10, 19, 27, 38, 48] \\
 & \ding{203}: [0, 9, 17, 30, 42]\\
 & \ding{204}: [18, 24, 31, 43, 45] \\
\midrule
\rowcolor{gray!10} \multicolumn{2}{l}{\textit{STAR-SMIE}~\cite{chen2024fine, zhou2023zero}:} \\
\multirow{3}{*}{\makecell[l]{NTU-60~\cite{shahroudy2016ntu} \\ (55/5 Split)}} & \ding{202}: [4, 19, 31, 47, 51] \\
 & \ding{203}: [12, 29, 32, 44, 59] \\ 
 & \ding{204}: [7, 20, 28, 39, 58] \\
\hdashline
\multirow{3}{*}{\makecell[l]{NTU-120~\cite{liu2019ntu}\\ (110/10 Split)}} & \ding{202}: [3, 18, 26, 38, 41, 60, 87, 99, 102, 110] \\
 & \ding{203}: [5, 12, 14, 15, 17, 42, 67, 82, 100, 119] \\
 & \ding{204}: [6, 20, 27, 33, 42, 55, 71, 97, 104, 118] \\
\hdashline
\multirow{3}{*}{\makecell[l]{PKU-MMD I~\cite{liu2017pku}\\ (46/5 Split)}} & \ding{202}: [3, 14, 29, 31, 49] \\
 & \ding{203}: [2, 15, 39, 41, 43] \\
 & \ding{204}: [4, 12, 16, 22, 36] \\
\bottomrule
\end{tabular}}
\end{table}


\section{Implementation Details}
\label{app:implementation_details}
Following prior works~\cite{gupta2021syntactically,chen2024fine}, we adopt Shift-GCN~\cite{cheng2020skeleton} as the skeleton encoder. For the text encoder, we use CLIP ViT-L/14@336px~\cite{radford2021learning}, consistent with~\cite{chen2024fine,chen2025neuron}. Both the encoder and decoder of the VAE are implemented as two-layer MLPs. For flow matching, we employ a single-layer DiT backbone~\cite{peebles2023scalable}. The training consists of two stages: the ``learning'' phase and the ``deciding'' phase, with 1,000 and 200 iterations, respectively. We use the AdamW optimizer with a weight decay of 0.01 and a learning rate of $1\times10^{-4}$. Logit-normal sampling~\cite{esser2024scaling} is applied to bias the training timesteps in flow matching. The batch size is set to 64. The hyperparameters $\lambda_{\mathrm{Align}}$ and $\lambda_{\mathrm{Flow}}$ are both set to 0.1, and the GZSL threshold $\gamma$ is fixed to 0.75. All experiments are implemented in PyTorch and conducted on a GeForce RTX 4090 Ti GPU. All ablation studies and qualitative analyses are used SynSE-based Shift-GCN features.


\section{Additional Performance Comparison}
\label{app:addtional_sota}


\noindent\textbf{Basic Split Benchmark Evaluation II}.
\label{app:sota_basic_ii}
We further compare our method with previous approaches under the cross-view and cross-setup evaluation protocols. Since SynSE does not provide pre-trained skeleton features for these settings, we employ the STAR-based 1s-Shift-GCN skeleton features for a fair performance comparison. As shown in Table~\ref{table:ntu60_xview_ntu120_xset} and Table~\ref{table:pkummd}, our method consistently outperforms prior works on both ZSL and GZSL metrics, demonstrating its strong robustness to variations in camera view and setup conditions.


\begin{table*}[!h]
\caption{Performance comparisons on the Xview task of NTU-60 and Xset task of NTU-120. The best and the second-best results are marked in \textcolor{red}{\textbf{Red}} and \textcolor{blue}{Blue}, respectively. All methods use STAR-based~\cite{chen2024fine} Shift-GCN skeleton features, as SynSE~\cite{gupta2021syntactically} does not provide Xview and Xset features. $^{\color{Orange}\boldsymbol{\ddagger}}$ denotes the two-stream fusion, while others are single-stream.}
\label{table:ntu60_xview_ntu120_xset}
\centering
\scalebox{0.75}{
\renewcommand{\arraystretch}{1}
\begin{tabular}{lccccccccccccccccc}
\toprule
\multirow{4}{*}{Method} &\multirow{4}{*}{Venue} & \multicolumn{8}{c}{NTU-60 (Xview)} & \multicolumn{8}{c}{NTU-120 (Xset)} \\
\cmidrule(r){3-10} \cmidrule(l){11-18} 
& & \multicolumn{4}{c}{55/5 Split} & \multicolumn{4}{c}{48/12 Split} & \multicolumn{4}{c}{110/10 Split} & \multicolumn{4}{c}{96/24 Split}\\
\cmidrule(r){3-6} \cmidrule(lr){7-10} \cmidrule(lr){11-14} \cmidrule(l){15-18} 
& & ZSL & \multicolumn{3}{c}{GZSL} &  ZSL & \multicolumn{3}{c}{GZSL} & ZSL & \multicolumn{3}{c}{GZSL} &  ZSL & \multicolumn{3}{c}{GZSL} \\
\cmidrule(r){3-3} \cmidrule(lr){4-6} \cmidrule(lr){7-7} \cmidrule(lr){8-10} \cmidrule(lr){11-11} \cmidrule(lr){12-14} \cmidrule(lr){15-15} \cmidrule(l){16-18}
& & $\mathcal{A}cc$ & $\mathcal{S}$ & $\mathcal{U}$ & $\mathcal{H}$ & $\mathcal{A}cc$ & $\mathcal{S}$ & $\mathcal{U}$ & $\mathcal{H}$ & $\mathcal{A}cc$ & $\mathcal{S}$ & $\mathcal{U}$ & $\mathcal{H}$ & $\mathcal{A}cc$ & $\mathcal{S}$ & $\mathcal{U}$ & $\mathcal{H}$ \\
\midrule
ReViSE~\cite{hubert2017learning} & ICCV 2017 & 54.4 & 25.8 & 29.3 & 27.4 & 17.2 & 34.2 & 16.4 & 22.1 & 30.2 & 4.0 & 23.7 & 6.8 & 13.5 & 2.6 & 3.4 & 2.9  \\
JPoSE~\cite{wray2019fine} & ICCV 2019 & 72.0 & 61.1 & 59.5 & 60.3 & 28.9 & 29.0 & 14.7 & 19.5 & 52.8 & 23.6 & 4.4 & 7.4 & 38.5 & 79.3 & 2.6 & 4.9  \\
CADA-VAE~\cite{schonfeld2019generalized} & CVPR 2019 & 75.1 & 65.7 & 56.1 & 60.5 & 32.9 & 49.7 & 25.9 & 34.0 & 52.5 & 46.0 & 44.5 & 45.2 & 38.7 & 47.6 & 26.8 & 34.3 \\
SynSE~\cite{gupta2021syntactically} & ICIP 2021 & 68.0 & 65.5 & 45.6 & 53.8 & 29.9 & 61.3 & 24.6 & 35.1 & 59.3 & 58.9 & 49.2 & 53.6 & 41.4 & 46.8 & 31.8 & 37.9 \\
SMIE~\cite{zhou2023zero} & ACMMM 2023 & 79.0 & - & - & - & 41.0 & - & - & - & 57.0 & - & - & - & 42.3 & - & - & -\\
STAR~\cite{chen2024fine} & ACMMM 2024 & 81.6 & \textcolor{blue}{71.9} & 70.3 & 71.1 & 42.5 & 66.2 & 37.5 & 47.9 & 65.3 & 59.3 & \textcolor{blue}{59.5} & 59.4 & 44.1 & 53.7 & 34.1 & 41.7\\
STAR++~\cite{chen2026star++} & TCSVT 2026 & 81.9 & 61.6 & 71.5 & 66.2 & 50.6 & 60.8 & 41.1 & 49.0 & 69.0 & 63.4 & 49.6 & 55.7 & 50.4 & 57.7 & 39.3 & 46.8 \\
Neuron$^{\color{Orange}\boldsymbol{\ddagger}}$~\cite{chen2025neuron} & CVPR 2025 & \textcolor{red}{\textbf{87.8}} & 70.6 & \textcolor{blue}{75.9} & \textcolor{blue}{73.2} & \textcolor{blue}{63.3} & \textcolor{blue}{65.3} & \textcolor{red}{\textbf{58.1}} & \textcolor{red}{\textbf{61.5}} & \textcolor{blue}{71.1} & \textcolor{red}{\textbf{67.5}} & 58.9 & \textcolor{blue}{62.9} & \textcolor{blue}{54.0} & \textcolor{red}{\textbf{67.0}} & \textcolor{blue}{44.9} & \textcolor{blue}{53.8} \\
\hdashline
\rowcolor{yellow!10} \textbf{\texttt{Flora} (Ours)} & This work & \textcolor{blue}{85.2} & \textcolor{red}{\textbf{82.7}} & \textcolor{red}{\textbf{76.5}} & \textcolor{red}{\textbf{79.5}}  & \textcolor{red}{\textbf{64.9}} &  \textcolor{red}{\textbf{75.4}} &  \textcolor{blue}{50.0} & \textcolor{blue}{60.1} & \textcolor{red}{\textbf{76.0}} & \textcolor{blue}{62.8} & \textcolor{red}{\textbf{65.1}} & \textcolor{red}{\textbf{63.9}} & \textcolor{red}{\textbf{63.7}} & \textcolor{blue}{55.4} & \textcolor{red}{\textbf{56.3}} & \textcolor{red}{\textbf{55.9}} \\
\bottomrule
\end{tabular}}
\end{table*}


\begin{table*}[!h]
\caption{Performance comparisons on PKU-MMD I dataset under the ZSL and GZSL setting. The best and the second-best results are marked in \textcolor{red}{\textbf{Red}} and \textcolor{blue}{Blue}, respectively. All methods use STAR-based~\cite{chen2024fine} Shift-GCN skeleton features, as SynSE~\cite{gupta2021syntactically} does not provide PKU-MMD features.}
\label{table:pkummd}
\centering
\scalebox{0.75}{
\renewcommand{\arraystretch}{1}
\begin{tabular}{lccccccccccccccccc}
\toprule
\multirow{4}{*}{Method} &\multirow{4}{*}{Venue} & \multicolumn{8}{c}{PKU-MMD I (Xsub)} & \multicolumn{8}{c}{PKU-MMD I (Xview)}\\
\cmidrule(r){3-10} \cmidrule(l){11-18} 
& & \multicolumn{4}{c}{46/5 Split} & \multicolumn{4}{c}{39/12 Split} & \multicolumn{4}{c}{46/5 Split} & \multicolumn{4}{c}{39/12 Split}\\
\cmidrule(r){3-6} \cmidrule(lr){7-10} \cmidrule(lr){11-14} \cmidrule(l){15-18} 
& & ZSL & \multicolumn{3}{c}{GZSL} &  ZSL & \multicolumn{3}{c}{GZSL} &  ZSL & \multicolumn{3}{c}{GZSL} &  ZSL & \multicolumn{3}{c}{GZSL} \\
\cmidrule(r){3-3} \cmidrule(lr){4-6} \cmidrule(lr){7-7} \cmidrule(lr){8-10} \cmidrule(lr){11-11} \cmidrule(lr){12-14} \cmidrule(lr){15-15} \cmidrule(l){16-18}
& & $\mathcal{A}cc$ & $\mathcal{S}$ & $\mathcal{U}$ & $\mathcal{H}$ & $\mathcal{A}cc$ & $\mathcal{S}$ & $\mathcal{U}$ & $\mathcal{H}$ & $\mathcal{A}cc$ & $\mathcal{S}$ & $\mathcal{U}$ & $\mathcal{H}$ & $\mathcal{A}cc$ & $\mathcal{S}$ & $\mathcal{U}$ & $\mathcal{H}$ \\
\midrule
ReViSE~\cite{hubert2017learning} & ICCV 2017 & 54.2 & 44.9 & 34.5 & 39.1 & 19.3 & 35.7 & 13.0 & 19.0 & 54.1 & 50.7 & 39.9 & 44.6 & 12.7 & 34.5 & 9.4 & 14.8 \\
JPoSE~\cite{wray2019fine} & ICCV 2019 & 57.4 & 67.0 & 43.0 & 52.4 & 27.0 & 64.8 & 26.5 & 37.6 & 53.1 & 72.9 & 42.5 & 53.7 & 22.8 & 57.6 & 20.2 & 29.9 \\
CADA-VAE~\cite{schonfeld2019generalized} & CVPR 2019 & 73.9 & \textcolor{blue}{76.2} & 51.8 & 61.7 & 33.7 & 69.0 & 29.3 & 41.1 & 74.5 & 79.9 & 61.5 & 69.5 & 29.5 & 62.4 & 28.3 & 39.0 \\
SynSE~\cite{gupta2021syntactically} & ICIP 2021 & 69.5 & \textcolor{red}{\textbf{77.8}} & 40.2 & 53.0 & 36.5 & 71.9 & 30.0 & 42.3 & 71.7 & 69.9 & 51.1 & 59.0 & 25.4 & 61.9 & 22.6 & 33.1 \\
SMIE~\cite{zhou2023zero} & ACMMM 2023 & 72.9 & - & - & - & 44.2 & - & - & - & 71.6 & - & - & - & 40.7 & - & - & -\\
STAR~\cite{chen2024fine} & ACMMM 2024 & 76.3 & 59.1 & 72.3 & 65.0 & 50.2 & \textcolor{blue}{72.7} & 44.7 & 55.4 & 75.4 & \textcolor{blue}{73.5} & \textcolor{red}{\textbf{72.2}} & \textcolor{blue}{72.8} & 50.5 & 69.8 & 47.5 & 56.5 \\
STAR++~\cite{chen2026star++} & TCSVT 2026 & \textcolor{blue}{77.1} & 69.9 & \textcolor{red}{\textbf{73.5}} & \textcolor{blue}{71.7} & \textcolor{blue}{55.4} & 71.2 & \textcolor{blue}{52.3} & \textcolor{blue}{60.3} & \textcolor{red}{\textbf{76.6}} & 72.2 & 69.0 & 70.6 & \textcolor{blue}{57.0} & \textcolor{blue}{75.1} & \textcolor{blue}{51.3} & \textcolor{blue}{60.9} \\
\hdashline
\rowcolor{yellow!10} \textbf{\texttt{Flora} (Ours)} & This work & \textcolor{red}{\textbf{79.1}} & \textcolor{red}{\textbf{76.0}} & \textcolor{blue}{65.9} & \textcolor{red}{\textbf{70.6}}  & \textcolor{red}{\textbf{55.4}} & \textcolor{red}{\textbf{74.5}} & \textcolor{red}{\textbf{52.3}}  & \textcolor{red}{\textbf{61.5}} & \textcolor{blue}{76.3}  &  \textcolor{red}{\textbf{76.0}} & \textcolor{blue}{71.4}  &  \textcolor{red}{\textbf{73.7}} & \textcolor{red}{\textbf{58.7}}  & \textcolor{red}{\textbf{77.2}}  & \textcolor{red}{\textbf{55.6}} & \textcolor{red}{\textbf{64.6}} \\
\bottomrule
\end{tabular}}
\end{table*}

\vspace{0.3em}
\noindent\textbf{Random Split Benchmark Evaluation II}.
\label{app:sota_random_ii}
We also compare our method with other approaches under the random split strategies proposed in STAR-SMIE~\cite{chen2024fine,zhou2023zero}, as shown in Table~\ref{table:random_split_star_based}. The results demonstrate that our method remains robust across different split strategies and consistently outperforms prior works. Notably, our method even surpasses the two-stream approaches, such as Neuron~\cite{chen2025neuron}, despite being a single-stream model without result stacking.


\begin{table}[!h]
\caption{Average performance comparison of three random seen-unseen splits on NTU-60 and PKU-MMD I datasets proposed by SMIE-STAR~\cite{zhou2023zero,chen2024fine} with Shift-GCN features. The best and the second-best results are marked in \textcolor{red}{\textbf{Red}} and \textcolor{blue}{Blue}, respectively. $^{\color{Orange}\boldsymbol{\ddagger}}$ denotes the two-stream fusion, while others are single-stream.}
\label{table:random_split_star_based}
\centering
\scalebox{0.75}{
\renewcommand{\arraystretch}{1.0}
\begin{tabular}{lcccc}
\toprule
\multirow{3}{*}{Method} & \multicolumn{2}{c}{NTU-60} &  \multicolumn{2}{c}{PKU-MMD I}\\
 & \multicolumn{2}{c}{55/5 (Xsub)} & \multicolumn{2}{c}{46/5 (Xsub)}\\
\cmidrule(r){2-3} \cmidrule(l){4-5} 
& ZSL & GZSL & ZSL & GZSL \\
\midrule
ReViSE~\cite{hubert2017learning} & 54.7 & 27.4 & 48.7 & 32.8  \\
JPoSE~\cite{wray2019fine} & 56.6 & 44.7 & 39.2 & 31.7 \\
CADA-VAE~\cite{schonfeld2019generalized} & 58.0 & 47.1 & 49.0 & 52.7 \\
SynSE~\cite{gupta2021syntactically} & 59.9 & 49.9 & 43.5 & 40.4 \\
SMIE~\cite{zhou2023zero} & 64.2 & - & 66.4 & - \\
STAR~\cite{chen2024fine} & 77.5 & 62.8 & 70.6 & 67.1 \\
STAR++~\cite{chen2026star++} & 79.5 & 62.4 & 73.6 & 68.3 \\
Neuron$^{\color{Orange}\boldsymbol{\ddagger}}$~\cite{chen2025neuron} & \textcolor{blue}{84.5} & \textcolor{blue}{71.2} & \textcolor{blue}{74.4} & \textcolor{red}{\textbf{69.2}} \\
\hdashline
\rowcolor{yellow!10} \textbf{\texttt{Flora} (Ours)} & \textcolor{red}{\textbf{85.1}} & \textcolor{red}{\textbf{71.7}} & \textcolor{red}{\textbf{76.5}} & \textcolor{blue}{68.4} \\
\bottomrule
\end{tabular}}
\end{table}

\vspace{0.3em}
\noindent\textbf{More Challenging Seen-Unseen Evaluation}.
\label{app:sota_challenge}
We further evaluate the efficiency of our method under reduced seen category priors, as shown in Table~\ref{table:ntu60_ntu120_40_30}. Even with fewer seen categories, our method still achieves competitive results compared with previous approaches, demonstrating its superior generalization capability. Notably, under the 30/30 split setting on NTU-60 (Xsub), our method shows a substantial improvement, highlighting its strong potential when trained with limited seen category priors.


\begin{table}[!h]
\caption{Performance comparisons on NTU-60 and NTU-120 with more challenging seen-unseen splits. The best and the second-best results are marked in \textcolor{red}{\textbf{Red}} and \textcolor{blue}{Blue}, respectively. All methods use our own pre-trained Shift-GCN skeleton features, as PURLS~\cite{zhu2024part} and~TDSM \cite{do2024tdsm} do not provide their pre-trained models.}
\label{table:ntu60_ntu120_40_30}
\centering
\scalebox{0.75}{
\renewcommand{\arraystretch}{1.0}
\begin{tabular}{lccccc}
\toprule
\multirow{2}{*}{Method} &\multirow{2}{*}{Venue} & \multicolumn{2}{c}{NTU-60 (Xsub)} & \multicolumn{2}{c}{NTU-120 (Xsub)}\\
\cmidrule(r){3-4} \cmidrule(l){5-6}
& & 40/20 & 30/30 & 80/40 & 60/60\\
\midrule
ReViSE~\cite{hubert2017learning} & ICCV 2017 & 24.3 & 14.8 & 19.5 & 8.3 \\
JPoSE~\cite{wray2019fine} & ICCV 2019 & 20.1 & 12.4 & 13.7 & 7.7 \\
CADA-VAE~\cite{schonfeld2019generalized} & CVPR 2019 & 16.2 & 11.5 & 10.6 & 5.7 \\
SynSE~\cite{gupta2021syntactically} & ICIP 2021 & 19.9 & 12.0 & 13.6 & 7.7 \\
PURLS~\cite{zhu2024part} & CVPR 2024 & 31.1 & 23.5 & 28.4 & 19.6 \\
ScoPLe~\cite{zhu2025semantic} & CVPR 2025 & \textcolor{blue}{32.0} & 18.2 & 25.3 & 15.7 \\
TDSM~\cite{do2024tdsm} & ICCV 2025 & \textcolor{red}{\textbf{36.1}} & \textcolor{blue}{25.9} & \textcolor{blue}{37.0} & \textcolor{blue}{27.2} \\
\hdashline
\rowcolor{yellow!10} \textbf{\texttt{Flora} (Ours)} & This work & 31.1 & \textcolor{red}{\textbf{35.7}} & \textcolor{red}{\textbf{40.1}} & \textcolor{red}{\textbf{29.0}} \\
\bottomrule
\end{tabular}}
\end{table}


\begin{table*}[!h]
\vspace{-10pt}
\caption{ZSL Comparison with other methods under low-shot training with SynSE-based~\cite{gupta2021syntactically} Shift-GCN features.}
\label{table:low_shot_training}
\vspace{-7pt}
\centering
\scalebox{0.75}{
\renewcommand{\arraystretch}{1.0}
\begin{tabular}{lcccccccccccccccc}
\toprule
\multirow{3}{*}{Method} & \multicolumn{8}{c}{NTU-60} & \multicolumn{8}{c}{NTU-120} \\
\cmidrule(r){2-9} \cmidrule(lr){10-17}
 & \multicolumn{4}{c}{55/5 (Xsub)} & \multicolumn{4}{c}{48/12 (Xsub)} & \multicolumn{4}{c}{110/10 (Xsub)} & \multicolumn{4}{c}{96/24 (Xsub)}\\
\cmidrule(r){2-5} \cmidrule(lr){6-9} \cmidrule(r){10-13} \cmidrule(lr){14-17}
& 1\% & 5\% & 10\% & 50\% & 1\% & 5\% & 10\% & 50\% & 1\% & 5\% & 10\% & 50\% & 1\% & 5\% & 10\% & 50\% \\
\midrule
ReViSE~\cite{hubert2017learning} & 51.0 & 58.0 & 58.0 & 56.9 & 9.8 & 11.1 & 15.6 & 15.6 & 14.8 & 14.8 & 23.6 & 20.3 & 5.2 & 7.3 & 7.8 & 8.5\\
JPoSE~\cite{wray2019fine} & 33.0 & 46.8 & 62.1 & 65.0 & 23.8 & 25.8 & 28.3 & 32.6 & 15.6 & 36.5 & 49.9 & 48.2 & 8.6 & 33.5 & 33.9 & 35.7 \\
CADA-VAE~\cite{schonfeld2019generalized} & 76.6 & 74.6 & 76.9 & 74.1 & 24.3 & 26.4 & 27.6 & 26.8 & 29.9 & 38.3 & 39.1 & 35.3 & 25.4 & 26.1 & 25.0 & 25.4 \\
SynSE~\cite{gupta2021syntactically} & 44.3 & 43.7 & 42.8 & 43.8 & 18.6 & 17.1 & 17.3 & 18.4 & 56.0 & 55.7 & 56.0 & 56.2 & 24.1 & 26.5 & 26.1 & 25.5 \\
SMIE~\cite{zhou2023zero} & 43.8 & 77.1 & 76.9 & 77.6 & 29.3 & 36.0 & 38.1 & 40.2 & 36.0 & 55.9 & 58.1 & 60.8 & 13.9 & 30.4 & 34.4 & 42.7 \\
SA-DAVE~\cite{li2024sa} & 21.1 & 45.1 & 60.4 & 81.3 & 18.7 & 16.5 & 20.0 & 30.4 & 14.4 & 28.5 & 40.9 & 55.6 & 9.5 & 12.3 & 21.9 & 34.7 \\
STAR~\cite{chen2024fine} & 40.6 & 75.5 & 77.0 & 79.1 & 11.6 & 32.9 & 35.3 & 37.5 & 18.9 & 41.8 & 46.8 & 53.2 & 8.6 & 31.3 & 33.0 & 34.5 \\
Neuron~\cite{chen2025neuron} & 47.7 & 76.9 & 79.4 & 81.5 & 20.7 & 34.1 & 45.3 & \textcolor{blue}{52.8} & 28.8 & 48.5 & 62.8 & 68.6 & 10.2 & 22.8 & 33.5 & 51.0 \\
FS-VAE~\cite{wu2025frequency} & \textcolor{blue}{79.3} & 79.3 & 79.4 & 78.9 & \textcolor{blue}{38.0} & 38.7 & 38.7 & 38.7 & \textcolor{blue}{72.7} & \textcolor{blue}{69.6} & \textcolor{blue}{69.6} & 70.2 & \textcolor{blue}{50.2} & \textcolor{blue}{49.7} & 47.9 & 48.5 \\
TDSM~\cite{do2024tdsm} & 78.5 & \textcolor{blue}{80.7} & \textcolor{blue}{82.3} & \textcolor{blue}{83.8} & 32.1 & \textcolor{blue}{49.2} & \textcolor{blue}{52.4} & 51.5 & 63.3 & 69.3 & 66.3 & \textcolor{blue}{71.9} & 43.9 & 49.1 & \textcolor{blue}{55.1} & \textcolor{blue}{59.7} \\
\hdashline
\rowcolor{yellow!10} \textbf{\texttt{Flora} (Ours)} & \textcolor{red}{\textbf{82.8}} & \textcolor{red}{\textbf{86.5}} & \textcolor{red}{\textbf{85.6}} & \textcolor{red}{\textbf{85.6}} & \textcolor{red}{\textbf{46.5}} & \textcolor{red}{\textbf{54.3}} & \textcolor{red}{\textbf{56.1}} & \textcolor{red}{\textbf{55.4}} & \textcolor{red}{\textbf{77.4}} & \textcolor{red}{\textbf{78.9}} & \textcolor{red}{\textbf{78.1}} & \textcolor{red}{\textbf{78.1}} & \textcolor{red}{\textbf{58.0}} & \textcolor{red}{\textbf{65.1}} & \textcolor{red}{\textbf{65.9}} & \textcolor{red}{\textbf{65.8}} \\
\bottomrule
\end{tabular}}
\vspace{-10pt}
\end{table*}

\vspace{0.3em}
\noindent\textbf{Per-instance Inference Time Comparison}.
\label{app:inference_time}
In Table~\ref{table:per_instance_inference_time}, we report the inference time as the number of candidate categories increases during per-instance inference. Notably, our method maintains an inference time of under one second even when matching against 1000 categories.

\begin{table}[!h]
\centering
\caption{Per-instance Inference Time}
\label{table:per_instance_inference_time}
\scalebox{0.85}{
\renewcommand{\arraystretch}{1.0}
\begin{tabular}{lcccccc}
\toprule
\# Cand. Classes & 5 & 10 & 50 & 100 & 500 & 1000 \\
\midrule
Time (ms) & 4.5 & 7.1 & 26.9 & 49.9 & 252.3 & 511.0 \\
\bottomrule
\end{tabular}}
\end{table}


\section{Additional Ablation Studies}
\label{app:additional_ablation_studies}


\noindent \textbf{Influence of Learning and Deciding Phases}.
\label{app:ab_learning_and_deciding_phases}
In Table~\ref{table:learning_deciding_analysis}, we analyze the contributions of the learning and deciding phases within the overall framework. The neighbor-aware mechanism plays a crucial role, indicating that high-quality cross-modal alignment serves as a cornerstone for zero-shot skeleton-based action recognition. Furthermore, when equipped with the open-flow classifier, the framework better preserves information during the recognition stage, leading to improved performance.

\begin{table}[!h]
\caption{Component analysis on learning and deciding phases. $^\dagger$baseline alignment (Sec.~\ref{sec:preliminaries}). $^\ddagger$similarity matching. $^\S$calibration strategy in~\cite{chen2024fine,chen2025neuron}.}
\label{table:learning_deciding_analysis}
\centering
\scalebox{0.75}{
\renewcommand{\arraystretch}{1.0}
\begin{tabular}{>{\hspace{1pt}}c<{\hspace{1pt}} >{\hspace{1pt}}c<{\hspace{1pt}} cc cc}
\toprule
\multirow{2}{*}{\makecell[c]{\cellcolor{\hl} Neighbor-aware\\\cellcolor{\hl} Semantic\\[-1ex]}} &
\multirow{2}{*}{\makecell[c]{\cellcolor{\hl} Open-form\\\cellcolor{\hl} Flow\\[-1ex]}}  &
\multicolumn{2}{c}{NTU-60 (48/12)} & \multicolumn{2}{c}{NTU-120 (110/10)}\\
\cmidrule(lr){3-4} \cmidrule(l){5-6} 
 & & ZSL & GZSL & ZSL & GZSL \\
\midrule
\xmark$^\dagger$ & \xmark$^\ddagger$ & 48.2 & 42.3$^\S$ & 71.1 & 55.8$^\S$ \\
\xmark$^\dagger$ & \cmark & 49.6 & 46.3 & 74.8 & 63.9 \\
\cmark & \xmark$^\ddagger$ & 56.7 & 51.6$^\S$ & 77.1 & 64.9$^\S$ \\
\rowcolor{yellow!10} \cmark & \cmark & \textbf{65.3} & \textbf{60.5} & \textbf{79.6} & \textbf{66.1} \\
\bottomrule
\end{tabular}}
\end{table}


\vspace{0.3em}
\noindent \textbf{Influence of Text Encoders}.
\label{app:ab_text}
As shown in Table~\ref{table:text_encoder}, the performance varies across different text encoders. Despite these discrepancies, the overall results remain strong under the ZSL setting. Interestingly, the best performance is not achieved with the most powerful model, \textit{i.e.}, ViT-H/14. For fairness and consistency with prior studies~\cite{chen2024fine, chen2025neuron}, we adopt the ViT-L/14@336px model in all experiments.

\begin{table}[!h]
\caption{Analysis of different text encoders on NTU-120 (Xsub).}
\label{table:text_encoder}
\centering
\scalebox{0.8}{
\renewcommand{\arraystretch}{1.0}
\begin{tabular}{lcccc}
\toprule
\multirow{2}{*}{Text Encoder}& \multicolumn{2}{c}{110/10 Split} & \multicolumn{2}{c}{96/24 Split}\\
\cmidrule(r){2-3} \cmidrule(l){4-5} 
& ZSL & GZSL & ZSL & GZSL \\
\midrule
ViT-B/32 & 77.1 & 61.2 & 62.1 & 47.9\\
ViT-B/16 & 77.7 & 63.9 & 62.5 & 46.4 \\
ViT-L/14 & 79.6 & 66.3 & 65.6 & 52.6 \\
\rowcolor{yellow!10} ViT-L/14@336px & 79.6 & 66.1 & 66.4 & 53.2 \\
ViT-H/14 & 73.5 & 62.6 & 66.3 & 52.0 \\
\bottomrule
\end{tabular}}
\end{table}

\begin{figure}[!h]
\centering
\includegraphics[width=0.99\linewidth]{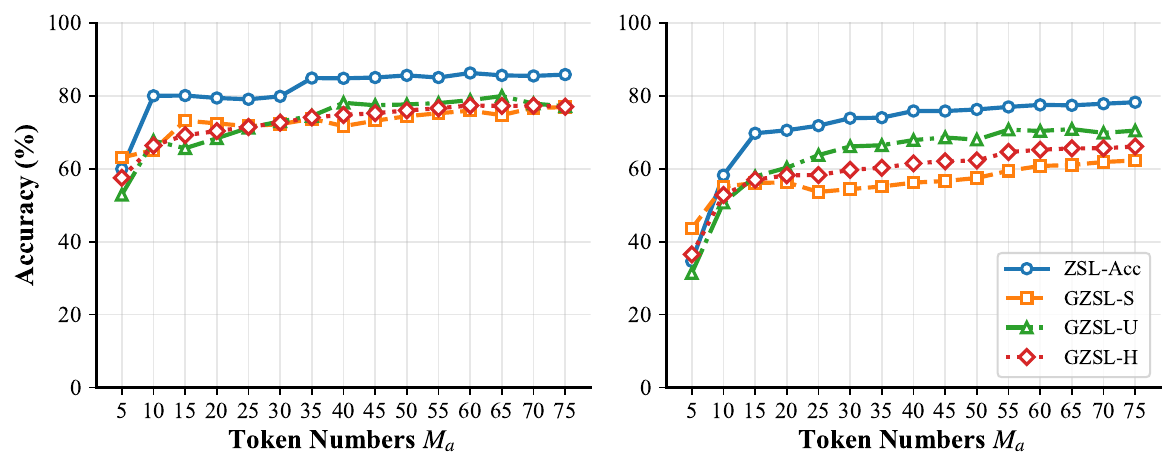} \\
\vspace{-5pt}
\begin{minipage}{0.99\linewidth}
\centering
\begin{tabular}{P{0.52}P{0.38}}
\footnotesize{(a) NTU-60 (55/5)} & 
\footnotesize{(b) NTU-120 (110/10)} \\
\end{tabular}
\end{minipage}
\vspace{-7pt}
\caption{Performance comparison on NTU-60 and NTU-120 under varying token numbers $M_a$.}
\label{fig:token_number}
\end{figure}

\vspace{0.3em}
\noindent \textbf{Influence of Token Numbers $M_a$}.
\label{app:ab_token}
As shown in Fig.~\ref{fig:token_number}, the performance of our method improves substantially as the number of tokens increases, and it gradually converges to a stable level when more tokens are involved. This trend suggests that enriching semantic representations contributes to more effective cross-modal alignment and that a sufficient number of tokens is essential to fully capture the semantic diversity required for robust performance.

\begin{figure*}[!h]
\centering
\includegraphics[width=0.99\linewidth]{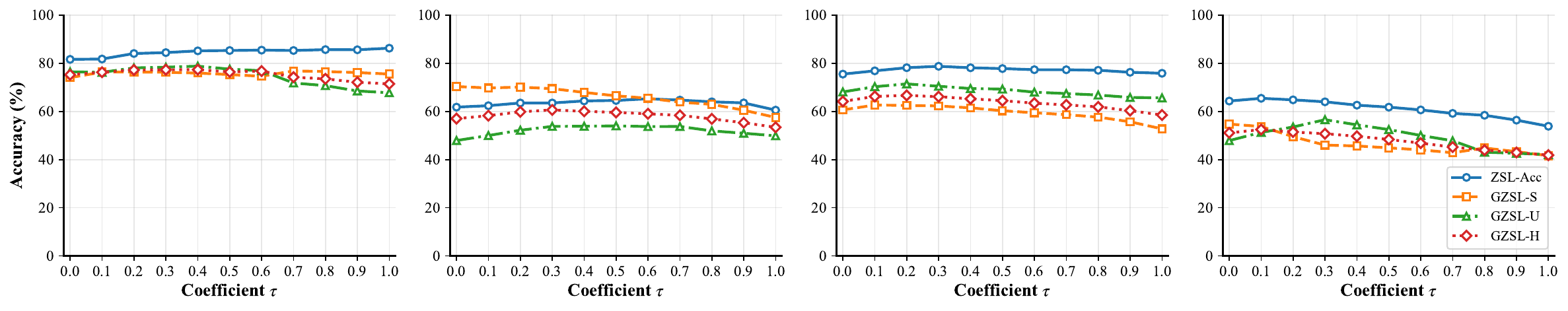} \\
\vspace{-5pt}
\begin{minipage}{0.99\linewidth}
\centering
\begin{tabular}{P{0.26}P{0.21}P{0.22}P{0.21}}
\footnotesize{(a) NTU-60 (55/5)} & 
\footnotesize{(b) NTU-60 (48/12)}  & 
\footnotesize{(c) NTU-120 (110/10)} & 
\footnotesize{(d) NTU-120 (96/24)} \\
\end{tabular}
\end{minipage}
\vspace{-7pt}
\caption{Performance comparison on NTU-60 and NTU-120 under varying coefficient $\tau$.}
\label{fig:coefficient_tau}
\end{figure*}

\vspace{0.3em}
\noindent \textbf{Influence of Coefficient $\tau$}.
\label{app:ab_coefficient_tau}
As illustrated in Fig.~\ref{fig:coefficient_tau}, both the harmonic accuracy and unseen performance first increase for smaller values of $\tau$ and then drop as $\tau$ varies. Overall, the performance trend stabilizes at a relatively high level, indicating that $\tau$ serves as a trade-off parameter that balances inter-class discriminability and the smoothness of the semantic space. Additionally, this coefficient is also robust to the selection of values.

\begin{figure*}[!h]
\centering
\includegraphics[width=0.99\linewidth]{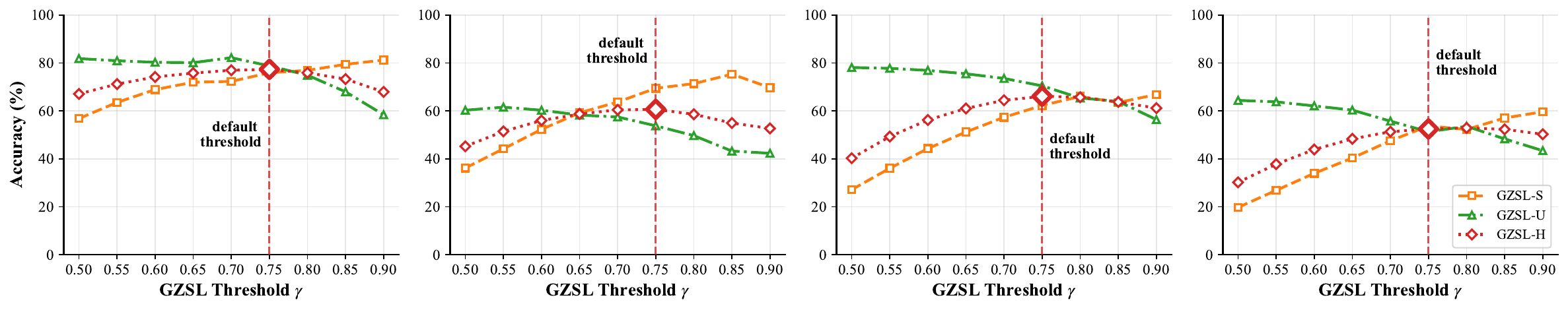} \\
\vspace{-5pt}
\begin{minipage}{0.99\linewidth}
\centering
\begin{tabular}{P{0.26}P{0.21}P{0.22}P{0.21}}
\footnotesize{(a) NTU-60 (55/5)} & 
\footnotesize{(b) NTU-60 (48/12)}  & 
\footnotesize{(c) NTU-120 (110/10)} & 
\footnotesize{(d) NTU-120 (96/24)} \\
\end{tabular}
\end{minipage}
\vspace{-7pt}
\caption{GZSL Performance comparison on NTU-60 and NTU-120 with varying predefined threshold $\gamma$.}
\label{fig:coefficient_gamma}
\end{figure*}


\vspace{0.3em}
\noindent \textbf{Influence of Threshold $\gamma$ in GZSL Prediction}.
\label{app:ab_threshold_gamma}
As shown in Fig.~\ref{fig:coefficient_gamma}, the GZSL performance is sensitive to the threshold $\gamma$, which controls whether a skeleton sample is classified as belonging to the seen or unseen domain. This behavior is expected, as $\gamma$ directly governs the gating mechanism in domain prediction. A higher threshold biases the model toward assigning skeleton samples to the seen domain, while a lower value favors the unseen domain.

\begin{figure*}[!h]
\centering
\includegraphics[width=0.99\linewidth]{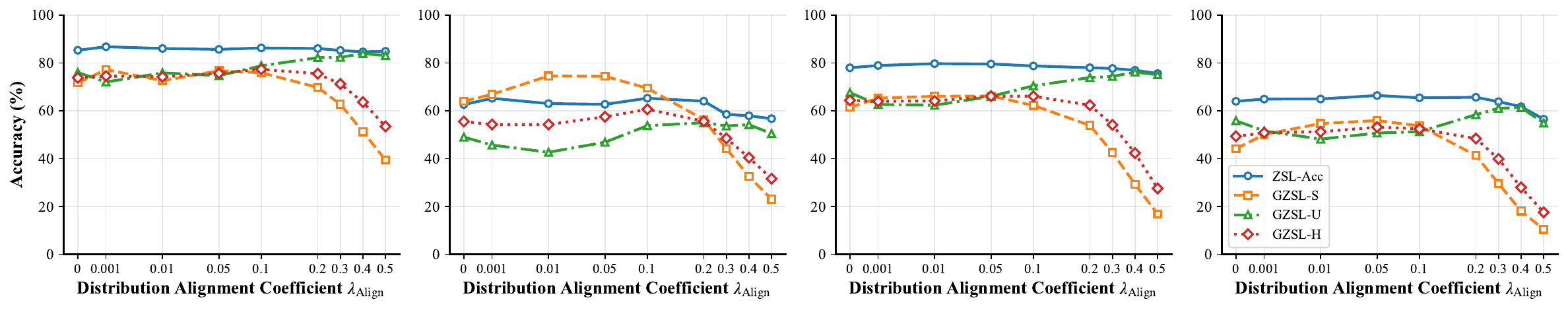} \\
\vspace{-5pt}
\begin{minipage}{0.99\linewidth}
\centering
\begin{tabular}{P{0.26}P{0.21}P{0.22}P{0.21}}
\footnotesize{(a) NTU-60 (55/5)} & 
\footnotesize{(b) NTU-60 (48/12)}  & 
\footnotesize{(c) NTU-120 (110/10)} & 
\footnotesize{(d) NTU-120 (96/24)} \\
\end{tabular}
\end{minipage}
\vspace{-7pt}
\caption{Performance comparison on NTU-60 and NTU-120 under various distribution alignment coefficient $\lambda_{\mathrm{Align}}$ in the learning phase.}
\label{fig:coefficient_alignment}
\end{figure*}

\vspace{0.3em}
\noindent \textbf{Influence of Distribution Alignment Coefficient $\lambda_{\mathbf{Align}}$ in Learning Phase}.
\label{app:ab_coefficient_align}
As shown in Fig.~\ref{fig:coefficient_alignment}, the performance exhibits an overall trend of increasing initially and then decreasing. When $\lambda_{\mathbf{Align}}$ exceeds 0.1, the performance drops sharply, particularly on the seen domains. This suggests that large $\lambda_{\mathbf{Align}}$ may cause the latent space to collapse, weakening the dominance of the reconstruction objectives.

\begin{figure*}[!h]
\centering
\includegraphics[width=0.99\linewidth]{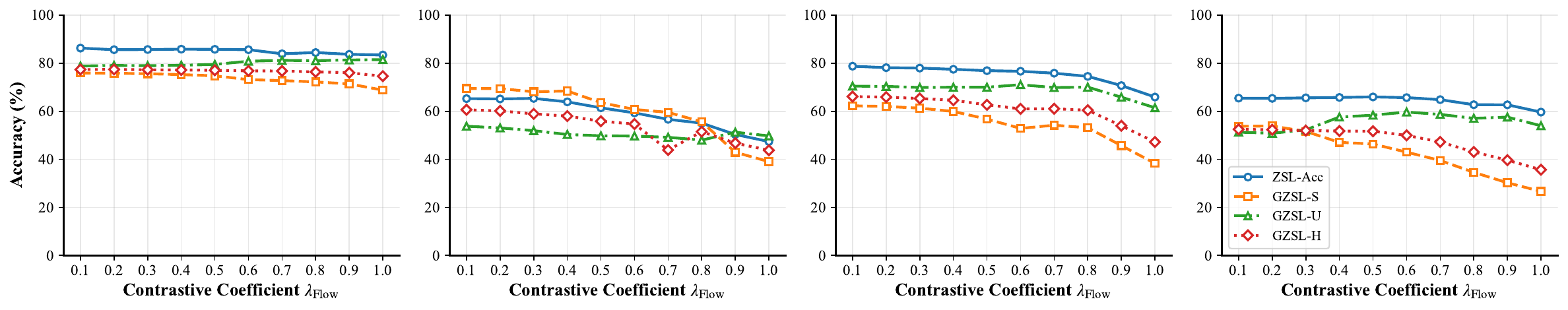} \\
\vspace{-5pt}
\begin{minipage}{0.99\linewidth}
\centering
\begin{tabular}{P{0.26}P{0.21}P{0.22}P{0.21}}
\footnotesize{(a) NTU-60 (55/5)} & 
\footnotesize{(b) NTU-60 (48/12)}  & 
\footnotesize{(c) NTU-120 (110/10)} & 
\footnotesize{(d) NTU-120 (96/24)} \\
\end{tabular}
\end{minipage}
\vspace{-7pt}
\caption{Performance comparison on NTU-60 and NTU-120 under different contrastive regularization coefficient $\lambda_{\mathrm{Flow}}$ in the deciding phase.}
\label{fig:coefficient_contrastive_fm}
\end{figure*}

\vspace{0.3em}
\noindent \textbf{Influence of Contrastive Regularization Coefficient $\lambda_{\mathrm{Flow}}$ in the Deciding Phase}.
\label{app:ab_coefficient_contrastive_fm}
As illustrated in Fig.~\ref{fig:coefficient_contrastive_fm}, the performance remains stable for smaller values of $\lambda_{\mathrm{Flow}}$ but declines as the coefficient increases. This suggests that mild contrastive regularization is beneficial for enhancing generalization, whereas an overly strong contrastive objective may hinder the classifier’s discriminative capability, especially for the seen categories.

\vspace{0.3em}
\noindent \textbf{Influence of Timestep Sampling Types in the Deciding Phase}.
\label{app:ab_timestep_sampling_types}
As shown in Table~\ref{table:timestep_sampling_types}, we compare the uniform-based and logit-based timestep sampling strategies. The results indicate that the choice of sampling type has minimal impact on the training of the flow classifier. In this work, we adopt the logit-based sampling strategy for consistency.


\begin{table}[!h]
\caption{Analysis of timestep sampling types in the deciding phase.}
\label{table:timestep_sampling_types}
\centering
\scalebox{0.8}{
\renewcommand{\arraystretch}{1.0}
\begin{tabular}{lcccc}
\toprule
\multirow{2}{*}{Types} & \multicolumn{2}{c}{110/10 Split} & \multicolumn{2}{c}{96/24 Split}\\
\cmidrule(r){2-3} \cmidrule(l){4-5} 
 & ZSL & GZSL & ZSL & GZSL \\
\midrule
Uniform-based & 78.6 & 65.6 & 65.7 & 52.0 \\
\hdashline
\rowcolor{yellow!10} \textbf{Logit-based (Ours)} & 79.6 & 66.1 & 66.4 & 53.2 \\
\bottomrule
\end{tabular}}
\end{table}


\vspace{0.3em}
\noindent \textbf{Influence of Flow Matching Backbone}.
\label{app:fm_backbone}
As shown in Table~\ref{table:fm_backbone}, we further investigate the performance of flow classifiers with different backbone architectures. Even with a simple two-layer MLP, the ZSL performance remains strong, showing only a slight degradation compared to a single-layer DiT block. This indicates that our flow classifier is largely independent of architectural complexity and can achieve competitive results with minimal network design.

\begin{table}[!h]
\caption{Analysis of flow matching backbone in deciding phase.}
\label{table:fm_backbone}
\centering
\scalebox{0.8}{
\renewcommand{\arraystretch}{1.0}
\begin{tabular}{lcccc}
\toprule
\multirow{2}{*}{Direction} & \multicolumn{2}{c}{110/10 Split} & \multicolumn{2}{c}{96/24 Split}\\
\cmidrule(r){2-3} \cmidrule(l){4-5} 
 & ZSL & GZSL & ZSL & GZSL \\
\midrule
MLP & 78.6 & 65.3 & 65.1 & 51.4 \\
\hdashline
\rowcolor{yellow!10} \textbf{DiT (Ours)} & 79.6 & 66.1 & 66.4 & 53.2 \\
\bottomrule
\end{tabular}}
\end{table}

\vspace{0.3em}
\noindent \textbf{Influence of Flow Directions in the Deciding Phase}.
\label{app:ab_flow_directions}
As shown in Table~\ref{table:flow_directions}, the choice of flow direction between skeleton and semantics has little effect on performance, since flow matching operates on interpolated vectors between the two modalities. In this work, we set the default flow direction from semantics to skeleton.

\begin{table}[!h]
\caption{Analysis of flow directions on the NTU-120 (Xsub) in the deciding phase.}
\label{table:flow_directions}
\centering
\scalebox{0.8}{
\renewcommand{\arraystretch}{1.0}
\begin{tabular}{lcccc}
\toprule
\multirow{2}{*}{Direction} & \multicolumn{2}{c}{110/10 Split} & \multicolumn{2}{c}{96/24 Split}\\
\cmidrule(r){2-3} \cmidrule(l){4-5} 
 & ZSL & GZSL & ZSL & GZSL \\
\midrule
Skeleton$\mathcal{N}_s\Rightarrow$Semantic$\mathcal{N}_a$ & 78.9 & 65.8 & 65.3 & 52.2 \\
\hdashline
\rowcolor{yellow!10} \textbf{Semantic$\mathcal{N}_a\Rightarrow$Skeleton$\mathcal{N}_s$ (Ours)} & 79.6 & 66.1 & 66.4 & 53.2 \\
\bottomrule
\end{tabular}}
\end{table}


\section{Additional Discussions}
\label{app:additional_discussions}

\begin{figure*}[t!]
\centering
\includegraphics[width=0.99\linewidth]{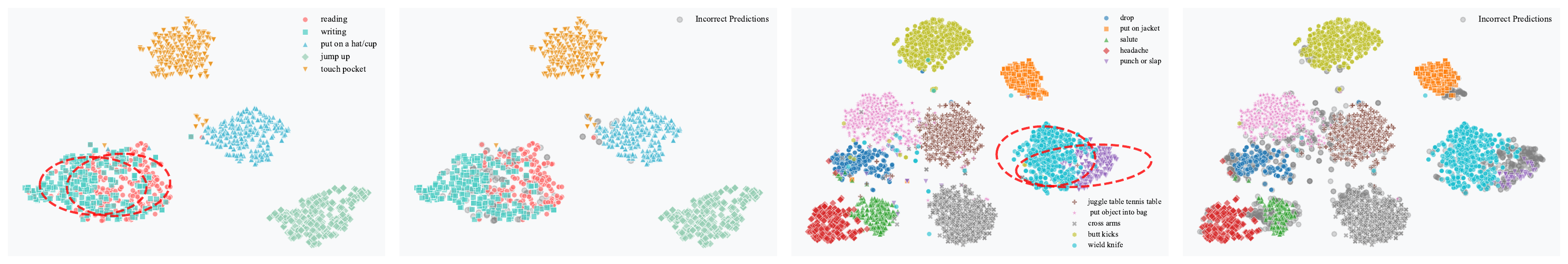} \\
\vspace{-5pt}
\begin{minipage}{0.99\linewidth}
\centering
\begin{tabular}{P{0.26}P{0.21}P{0.22}P{0.21}}
\footnotesize{(a) NTU-60 (55/5)} & 
\footnotesize{(b) NTU-60 (55/5)}  & 
\footnotesize{(c) NTU-120 (110/10)} & 
\footnotesize{(d) NTU-120 (110/10)} \\
\end{tabular}
\end{minipage}
\vspace{-7pt}
\caption{t-SNE visualization on NTU-60 (55/5) and NTU-120 (110/10).}
\label{fig:tsne}
\vspace{-10pt}
\end{figure*}

\noindent \textbf{Skeleton Perspective}.
\label{app:ad_skeleton}
We summarize and discuss zero-shot skeleton recognition from the perspective of skeleton as follows:
\begin{itemize}
    \item \textit{Low-shot Training Samples}. As shown in Table~\ref{table:low_shot_training}, our experiments demonstrate the promising potential of zero-shot skeleton action recognition toward more efficient learning. This observation motivates us not only to focus on limited seen categories but also to explore learning from limited samples. Such a direction suggests that it is feasible to build an intelligent system with strong generalizability and robustness, even when trained with a small number of samples from a few categories.
    \item \textit{Representation Quality}. Another key challenge lies in the limited information contained in skeleton data. For instance, a single joint is often used to represent an entire hand, which leads to overlapping skeleton features across similar actions (Fig.~\ref{fig:tsne}), such as reading and writing. This overlap makes it difficult to separate features from different categories, particularly for unseen ones, since their priors are unavailable during training. Therefore, incorporating finer-grained skeleton representations—such as increasing the number of joints to capture more detailed motion—may be a promising direction for advancing skeleton-based community, beyond the zero-shot setting.
\end{itemize}

\vspace{0.3em}
\noindent \textbf{Semantic Perspective}.
\label{app:ad_semantic}
We further discuss it from the perspective of semantics as follows:
\begin{itemize}
    \item \textit{Skeleton-specific Semantics}. Current semantics are typically action-specific, whether derived from hand-crafted labels or LLM-generated descriptions, and thus are not inherently aligned with the nature of skeleton representations. For instance, the semantics of “pick up” share little linguistic similarity with “put on a shoe”, yet their skeleton sequences are highly similar, as both involve a squatting motion. This discrepancy, where distant semantics correspond to highly similar skeletal patterns, leads to cross-modal structural inconsistency prior to alignment. On such a fragile foundation, building a reliable semantic–skeleton alignment becomes inherently difficult. Therefore, designing skeleton-structural semantics that are consistent with the physical motion patterns is crucial, though largely overlooked in existing research.
    \item \textit{Semantic Diversity}. Action descriptions can vary significantly across observation viewpoints or subjects with different body shapes. Incorporating diverse semantics that account for these variations is essential for achieving robust alignment. A promising direction is to leverage sample-level semantics for alignment. It effectively reframes the recognition task as a zero-shot captioning problem, where the model learns to describe actions through semantically grounded understanding rather than rigid label matching. In this setting, we believe \texttt{\textbf{Flora}} can play a vital role.
\end{itemize}

\vspace{0.3em}
\noindent \textbf{Algorithm Perspective}.
\begin{itemize}
    \item \textit{Alignment}. Similar to how a limited set of pixels with diverse combinations can generate an infinite number of images and promote zero-shot learning in various domains, exploring the compositionality of skeletal primitives (such as fixed joint groups or joint motion velocities) is equally important. A finite number of primitives with different variations can represent an unlimited range of actions. In contrast to existing paradigms that rely solely on pre-extracted skeleton features for alignment, developing a continual skeleton composition framework can enable cross-modal alignment at a more fundamental level, thereby enhancing the performance of zero-shot skeleton-related tasks.
    \item \textit{Task}. Beyond recognition, building a skeleton-based foundation model capable of handling various tasks, including skeleton captioning and generation, under zero-shot settings is a promising direction. Our proposed \texttt{\textbf{Flora}} framework provides a paradigm for these advancements by establishing a dynamic flow-based pathway between skeletons and semantics, effectively bridging the gap between perception and understanding.
\end{itemize}
\label{app:ad_algorithm}

\end{document}